\documentclass[final,12pt]{elsarticle}
\usepackage{amssymb}
\usepackage{svg}
\usepackage{amsmath}
\usepackage{multirow}
\usepackage{booktabs}
\usepackage{float}
\usepackage{array}
\usepackage{graphicx}    % 用于插入图片
\usepackage{subcaption}  % 提供 subfigure 环境
\usepackage{hyperref}
\usepackage{algorithm}
\usepackage{algpseudocode}
\usepackage{amsmath} % 用于数学公式
\usepackage{xcolor}
\usepackage{titlesec}

\titleformat{\paragraph}[block] % ← 关键：改为 [block] 实现换行
  {\normalfont\normalsize}      % ← 关键：去掉 \bfseries，不加粗！
  {\theparagraph}
  {1em}
  {}

\titlespacing*{\paragraph}
  {0pt}                        % 左缩进
  {1ex}                        % 段前间距（可调）
  {1em}                        % 段后间距 —— 控制标题与正文之间的空隙

\renewcommand{\theparagraph}{\thesubsubsection.\arabic{paragraph}}

\newcommand{\subsubsubsection}[1]{\paragraph{#1}} % 现在用 \subsubsubsection 就等价于 \paragraph

\usepackage{xcolor}
\hypersetup{
    pdfborder={0 0 0} % 禁用所有链接边框
}
\journal{Neurocomputing}

\begin{document}

\begin{frontmatter}



\title{SEAL: Self-Evolving Agentic Learning for Conversational Question Answering over Knowledge Graphs} %% Article title

% ====== 修复脚注符号冲突 ======
\makeatletter
\def\@fnsymbol#1{\ensuremath{\ifcase#1\or *\or \dagger\or \ddagger\or
   \mathsection\or \mathparagraph\or \|\or **\or \dagger\dagger
   \or \ddagger\ddagger \else\@ctrerr\fi}}
\makeatother

\author[label1]{Hao Wang}
\author[label2]{Jialun Zhong}
\author[label2]{Changcheng Wang}
\author[label3]{Zhujun Nie\corref{intern}}
\author[label5]{Zheng Li}
\author[label1]{Shunyu Yao}
% \author[label4,label2,cor]{Yanzeng Li}
% \author[label1,cor]{Xinchi Li}

\author[label4,label2]{Yanzeng Li\corref{cor}}
\author[label1]{Xinchi Li\corref{cor}}

\affiliation[label1]{organization={Institute of Big Data and Artificial Intelligence, China Telecom Research Institute},
            city={Beijing},
            postcode={102209}, 
            country={China}}

\affiliation[label2]{organization={Wangxuan Institute of Computer Technology, Peking University},
            city={Beijing},
            postcode={100871}, 
            country={China}}

\affiliation[label3]{organization={School of Artificial Intelligence, China University of Geosciences (Beijing)},
            city={Beijing},
            postcode={100083}, 
            country={China}}

\affiliation[label5]{organization={Center for Cognition and Neuroergonomics, State Key Laboratory of Cognitive Neuroscience and Learning, Beijing Normal University},
            city={Zhuhai},
            postcode={519087}, 
            state={Guangdong},
            country={China}}

\affiliation[label4]{organization={Institute of Artificial Intelligence and Future Networks, Beijing Normal University},%Department and Organization
            city={Zhuhai},
            postcode={519087}, 
            state={Guangdong},
            country={China}}

\cortext[cor]{Corresponding Author}
\cortext[intern]{This work was done during the internship at China Telecom Research Institute.}

%% Abstract
\begin{abstract}

Knowledge-based conversational question answering (KBCQA) confronts persistent challenges in resolving coreference, modeling contextual dependencies, and executing complex logical reasoning.
Existing approaches, whether end-to-end semantic parsing or stepwise agent-based reasoning, often suffer from inaccuracies and prohibitive computational costs, particularly when processing intricate queries over large knowledge graphs. Specifically, large language models (LLMs) tend to generate syntactically invalid or semantically misaligned logical forms for complex multi-hop or aggregation queries. Meanwhile, conventional entity-relation linking methods face an exponentially growing candidate space, leading to high latency and error propagation.
To address these limitations, we introduce SEAL, a novel two-stage semantic parsing framework grounded in self-evolving agentic learning.
In the first stage, an LLM extracts a minimal S-expression core that captures the essential semantics of the input query. This core is then refined by an agentic calibration module, which corrects syntactic inconsistencies and aligns entities and relations precisely with the underlying knowledge graph. The second stage employs template-based completion, guided by question-type prediction and placeholder instantiation, to construct a fully executable S-expression. This decomposition not only simplifies logical form generation but also significantly enhances structural fidelity and linking efficiency.
Crucially, SEAL incorporates a self-evolving mechanism that integrates local and global memory with a reflection module, enabling continuous adaptation from dialog history and execution feedback without explicit retraining. Extensive experiments on the SPICE benchmark demonstrate that SEAL achieves state-of-the-art performance, especially in multi-hop reasoning, comparison, and aggregation tasks. The results validate notable gains in both structural accuracy and computational efficiency, underscoring the framework's capacity for robust and scalable conversational reasoning.
\end{abstract}


\end{frontmatter}
\section{Introduction}

A Knowledge Graph (KG) is a structured representation of knowledge, typically organized as triples (head entity, relation, tail entity) to encode factual information~\cite{SurveyKnowledgeGraphs}. In recent years, KGs have gained widespread attention in both academia and industry~\cite{Zou_2020, bader2020knowledge}. Knowledge-based Question Answering (KBQA) systems are designed to query these structured KGs, using reasoning to provide accurate answers to natural language questions~\cite{lan2022complex, diefenbach2018core}. Among KBQA methods, Semantic Parsing (SP) based approaches translate questions into structured queries (e.g., SPARQL, Cypher, etc.) for execution against the KG, offering strong interpretability and high efficiency~\cite{feng2025rgr, nam2025semantic}. These systems are widely applied in fields such as healthcare and business, significantly reducing the technical threshold for accessing complex knowledge systems. Knowledge-based conversational QA (KBCQA) extends this paradigm to multi-turn interactive scenarios, requiring the system to conduct continuous reasoning and to address dialog understanding challenges such as coreference resolution~\cite{liu2023brief, lee2017end}. For this task, SP remains a mainstream approach, where the goal is to convert contextual natural language queries into executable logical forms. With the emergence of large language models (LLMs)~\cite{minaee2024large, zhao2026survey}, SP increasingly leverages their advanced language understanding capabilities~\cite{zhang2023two,zhan2024progressive,wang2025question}, primarily through two paradigms: end-to-end logical form generation and agent-based stepwise construction.

While LLMs offer significant opportunities for SP-based KBQA, and KBCQA tasks, current methods face substantial limitations in handling structurally complex questions~\cite{xuainterpretable}. Specifically, generated logical forms often fail to fully capture semantic intent in scenarios requiring multi-hop reasoning, comparison, or aggregation operations~\cite{kamath2018survey, wei2023semantic, du2021cogkr}. This limitation is particularly evident in complex logical reasoning, where LLMs tend to focus on surface-level concepts while overlooking critical structural constraints imposed by the knowledge graph .Furthermore, the entity and relation linking process suffers from an expansive candidate space due to linguistic ambiguity~\cite{wen2025clear, li2023cord}, leading to exponential growth in possible combinations and high computational overhead. This issue directly impacts reasoning generalization, LLMs often generate plausible but semantically invalid forms that ignore domain-specific validity constraints.These challenges hinder the scalability of SP-based KBQA systems, and are further exacerbated in the KBCQA setting, where the system must also manage dialog history to resolve coreferences and maintain contextual coherence.In particular,coreference resolution remains a major bottleneck that if but without aligning the resolved entity with its attributes in the knowledge graph , the final answer can still be inconsistent or incorrect.

\begin{figure}[ht]
    \centering
    \includegraphics[width=1\linewidth]{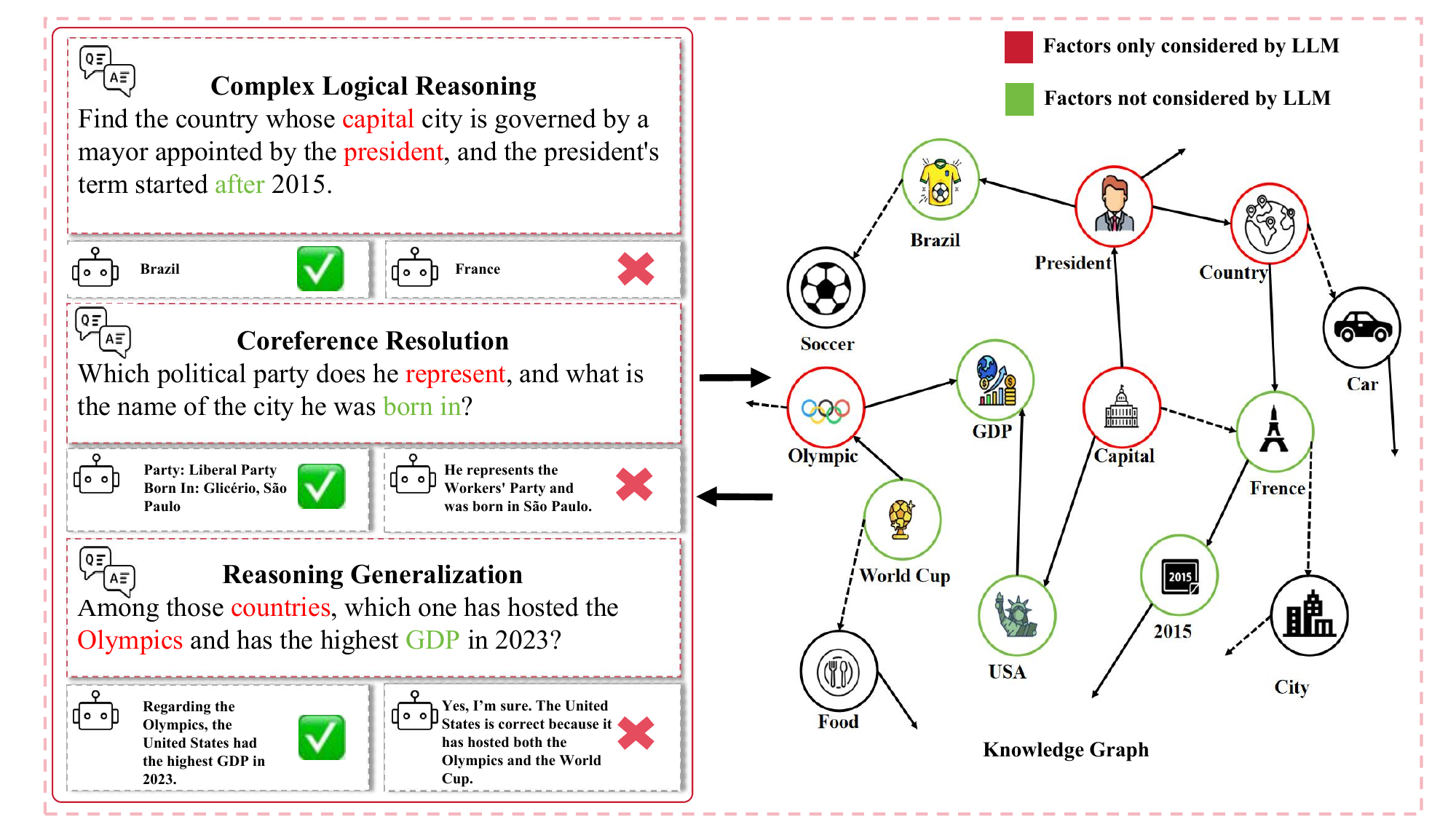}
    \caption{The challenges of leveraging LLMs in KBCQA.}
    \label{figure:knowledge graph}
\end{figure}

A key research problem in KBCQA is how to leverage LLMs to address the challenges of generating complex logical structures and the high computational cost of entity linking, as shown in Figure~\ref{figure:knowledge graph}. To this end, this article introduces Self-Evolving Agentic Learning (SEAL), a two-stage SP framework. SEAL leverages S-expressions, a structured logical form. This clear and readable structure is particularly advantageous for representing the complex and discrete operations required by the KBCQA task~\cite{nam2025semantic,feng2025rgr}.

In the first stage, the LLM generates a preliminary S-expression core, which is then semantically calibrated by an agent to correct structural errors. In the second stage, the LLM completes the logical structure by integrating the validated core with predefined templates, producing an accurate and executable S-expression. Crucially, SEAL incorporates a self-evolving mechanism that establishes a continuously learning agent through the synergy of local memory, global memory, and a reflection module. This mechanism enables the system to adaptively learn from successful past dialogs and execution outcomes, transforming global memory from static storage into a dynamically updateable knowledge base without explicit retraining. This approach effectively combines the semantic understanding of LLMs with the structural rigor of templates, improving the accuracy of complex query generation in KBCQA while maintaining high efficiency.

We summarize the contributions of this paper as follows:
\begin{itemize}
    \item We introduce the concept of a minimal S-expression core to represent the essential semantics of a query. This core is calibrated by an agent for syntactic correctness and knowledge graph alignment, forming a robust foundation for the final query construction.
    \item We propose SEAL, a two-stage agentic learning framework for the KBCQA task, in which innovations are decomposition of SP into S-expression core extraction followed by agent-driven calibration and template-based composition, significantly enhancing structural accuracy and computational efficiency in complex reasoning.
    \item We design a self-evolving mechanism that enables continuous performance enhancement through dynamic memory updates and reflection, allowing the system to adapt to novel expressions in real-world dialogs without retraining.
    \item Extensive experiments on the SPICE benchmark show that our method achieves state-of-the-art results, particularly in complex reasoning tasks, with the self-evolving mechanism demonstrating significant performance improvements as dialog progresses. The results validate significant improvement in both structural accuracy and efficiency.
\end{itemize}

The article is structured as follows:
Section~\ref{sec:rw} reviews related works on SP in KBQA, KCBQA, and LLM-based agents. Section~\ref{sec:preli} provides preliminaries on knowledge graphs. Section~\ref{sec:method} details our proposed method, including the reasoning, memory, and reflection modules. Section~\ref{sec:exp} presents the experimental setup, datasets, metrics, baselines, and main results.  Section~\ref{sec:conclu} concludes the article.
\section{Related Work}
\label{sec:rw}

In this section, we introduce  work related to our research, covering SP-based KBQA, KBCQA, and LLM-based Agents.

\subsection{Semantic Parsing in KBQA}
Early research on KBQA focused on SP to translate natural language questions into structured queries, ensuring interpretability and logical reasoning. In an early work, a staged framework \cite{yihSemanticParsingStaged2015} decomposes query generation into entity recognition, inference chain construction, and constraint aggregation, providing a foundation for natural language to graph query generation. Graph embeddings and constraint-based path control \cite{baoConstraintBasedQuestionAnswering2016} reduce the search space in multi-hop reasoning, emphasizing structural efficiency. Query construction as a state-transition process \cite{huStatetransitionFrameworkAnswer2018} employs node identification, connection, merging, and folding operations for dynamic semantic dependencies. With the rise of LLMs, KBQA shifted towards data-driven paradigms, with few-shot prompting \cite{liFewshotIncontextLearning2023} reducing reliance on annotated data and enhancing adaptability. Query generation aligned with code synthesis paradigms \cite{nieCodeStyleInContextLearning2024} leverages structured programming syntax. A generation and retrieval strategy \cite{luoChatKBQAGeneratethenRetrieveFramework2024} improves multi-hop reasoning, while external knowledge retrieval \cite{xuHarnessingLargeLanguage2025} enriches logic forms, addressing knowledge incompleteness. Agentic approaches facilitate dynamic query construction, with symbolic agents \cite{guDontGenerateDiscriminate2023a} improving the precision of relational inference. Step-by-step reasoning within a thought action cycle \cite{xiongInteractiveKBQAMultiTurnInteractions2024} enables progressive refinement. A mechanism of observation, action, and reflection \cite{sunODAObservationDrivenAgent2024} enhances robustness. Planning, retrieval, and reasoning \cite{luoReasoningGraphsFaithful2024} support structured query generation on heterogeneous graphs.

\subsection{Knowledge-based Conversational QA}

KBCQA presents greater challenges than single-turn tasks, particularly in modeling dialog history and integrating heterogeneous knowledge. A comprehensive survey \cite{vadhavanaConversationalQuestionAnswering2024} traces the evolution of CQA, identifying context modeling issues. The Complex Sequential QA (CSQA) Dataset \cite{ComplexSequentialQuestion} incorporates dialog history into complex reasoning, while the CONVINSE framework and the ConvMix dataset \cite{christmannConversationalQuestionAnswering2023} enable multi-source reasoning. Datasets with SPARQL annotations support consistent logic parsing across turns.We choose the S-expression to represent the logical form of questions~\cite{perez-beltrachiniSemanticParsingConversational2023a}. Proposed by Gu et al.~\cite{guIIDThreeLevels2021}, S-expressions are a Lisp-based format that uses functions to express logical relationships, widely applied in recent works such as KB-BINDER~\cite{liFewshotIncontextLearning2023}, KB-Coder~\cite{nieCodeStyleInContextLearning2024}, and Pangu~\cite{guDontGenerateDiscriminate2023a}.  Ambiguity resolution, including pronoun coreference and ellipsis, remains a challenge, with traditional and neural approaches \cite{zhangBriefSurveyComparative2021} identifying limitations for rare entities. LLM-driven disambiguation strategies \cite{DisambiguationConversationalQuestion} address ellipsis and semantic ambiguity, while CoQA, SQuAD 2.0, and QuAC \cite{yatskarQualitativeComparisonCoQA2019} reveal deficiencies in dynamic dialog. Question rewriting transforms context-dependent questions into self-contained forms, with a two-stage pipeline \cite{vakulenkoQuestionRewritingConversational2020} improving retrieval. Reinforcement learning \cite{chenReinforcedQuestionRewriting2022} optimizes rewriting via QA feedback. Question rewriting variants \cite{raposoQuestionRewritingAssessing2022} enhance context representation. The REIGN framework \cite{kaiserRobustTrainingConversational2024} uses data augmentation and reinforcement learning. CornNet \cite{liuConversationalQuestionAnswering2024} integrates LLM-based rewriting with teacher-student architectures. Dialog history modeling integrates explicit memory and entity tracking, with Dialog-to-Action \cite{guoDialogtoActionConversationalQuestion2018a} resolving ellipsis via memory management. Graph neural networks \cite{DBLP:conf/emnlp/JainL23} encode evolving subgraphs. LLMs with dynamic memory \cite{jainIntegratingLargeLanguage2024} synthesize diverse evidence.  Reinforcement learning \cite{xuReinforcementLearningConstraints2024} tracks entities across multi-hop graphs. The Adaptive Context Management framework \cite{pereraAdaptiveContextManagement2025} adjusts context windows for relevance. The KaFSP framework \cite{liKaFSPKnowledgeAwareFuzzy2022} integrates fuzzy reasoning, and LLM scalability \cite{guoDialogtoActionConversationalQuestion2018a} confirms effectiveness of few-shot prompting and fine-tuning.

\subsection{LLM-based Agents}
Leveraging vast pre-training on expansive corpora and subsequent instruction fine-tuning \cite{DBLP:conf/nips/Ouyang0JAWMZASR22} across a diverse array of tasks, LLMs exhibit exceptional capabilities in representation, reasoning, and generation, facilitating their application across a broad spectrum of language-mediated challenges. This foundation has spurred significant interest in LLM-based agents~\cite{DBLP:journals/corr/abs-2503-21460}, which have gained widespread attention \cite{DBLP:journals/corr/abs-2503-11733,li2026community,tao2026llm} due to their intellectual proficiency, driving advancements in adaptive task execution. These agents integrate LLMs into sophisticated, human-like cognitive frameworks, incorporating key components such as perception \cite{DBLP:conf/icml/DriessXSLCIWTVY23,DBLP:conf/iclr/ZhengLFL24}, strategic planning \cite{DBLP:journals/corr/abs-2402-02716,DBLP:conf/acl/WeiZHXPL25}, and actionable execution \cite{DBLP:conf/hri/KimLM24,DBLP:conf/nips/LiZWWZSGLLZLL0M24}, enabling robust adaptation to dynamic and complex environments \cite{DBLP:conf/iclr/0036YZXLL0DMYZ024}.To enhance their capability to invoke tools within complex reasoning, ChatCoT \cite{chenChatCoTToolAugmentedChainofThought2023} models Chain-of-Thought (CoT) reasoning as a multi-round dialogue, enabling the unified integration of CoT and tool operation. Furthermore, specialized modules tailored for long-term tasks enhance their efficacy: memory systems, encompassing symbolic \cite{DBLP:journals/corr/abs-2306-03901} and textual \cite{DBLP:journals/corr/abs-2311-08719} summaries of past interactions \cite{DBLP:journals/tmlr/WangX0MXZFA24}, support sustained contextual awareness, while reflection mechanisms \cite{DBLP:journals/corr/abs-2306-03901} foster self-evolution and adaptability.These mechanisms of self-correction and reflection are further explored, where Self-Refine \cite{madaanSelfRefineIterativeRefinement2023} employs iterative self-feedback to optimize generative outputs, and Reflexion \cite{shinnReflexionLanguageAgents2023} utilizes verbal reflections as episodic memory to guide subsequent actions. Our approach aligns with this paradigm, harnessing these strengths to enable continuous performance enhancement in response to streaming data, without the need for retraining.
\section{Preliminaries}
\label{sec:preli}
\subsection{Knowledge Graph} Let ${E}$, ${R}$ denote the sets of entities and relations respectively. A knowledge graph can be represented as \( {G} = ({E}, {R}, {T}) \) , where \( {T} \subseteq {E} \times {R} \times {E} \) is the set of facts stored in the KG. A fact in ${T}$ can be represented as a triple \( (e_h, r, e_t)\), indicating that a directed relation \( r \in R \) holds between a head entity \( e_h \in E \) and a tail entity \( e_t \in E \).
%%%ConvQA
\subsection{KG-based conversational QA} For the KG-based conversational QA task, a dialogue \( d \in {D} \) consists of sequential turns of questions and answers \( d = (q_1, a_1, q_2, a_2, \dots, q_n, a_n) \).
The types of answers $a_i$ include sets of entities, Boolean values, aggregation quantities, etc., and need to be inferred based on the question $q_i$, dialogue history $H_i = (q_1, a_1, \dots, q_{i-1}, a_{i-1})$, and the given knowledge graph \( G \). This process is represented in Equation~\ref{eq:transform}:
\begin{equation}
H_i \times q_i \times G \rightarrow a_i
\label{eq:transform}
\end{equation}

Moreover, some recent work parse the natural language question $q_i$ and map it onto an executable logic form \( f_i \in F \) (e.g., SPARQL and S-expressions) on the KG, leading to an explicit reasoning process.

\section{Method}
\label{sec:method}
\subsection{Overview}
We propose a novel SP approach for KBCQA that uses LLMs to directly generate S-expressions. However, LLM outputs often contain ungrounded surface forms, and conventional entity and relation linking methods that rely on large candidate sets are computationally expensive. 
To enable efficient and accurate parsing, we introduce a lightweight calibration strategy that performs syntax correction and single-candidate KG alignment. 

% We present a novel SP approach for KBCQA tasks~\cite{perez-beltrachiniSemanticParsingConversational2023a}, leveraging LLMs for SP directly into S-expressions. However, traditional methods suffer from high computational costs due to large-scale entity and relation linking. To mitigate this issue and ensure efficient parsing, we propose a lightweight expression calibration strategy.

\begin{table}[H]
\centering % Centers the table
\caption{S-expression Functions}
\label{tab:s-expr-functions}
\resizebox{\textwidth}{!}{
\begin{tabular}{|>{\ttfamily}l|>{\ttfamily}l|p{4cm}|} 
\hline
\textrm{\textbf{Function}} & \textrm{\textbf{Return Type}} & \textrm{\textbf{Description}} \\
\hline
\textbf{(JOIN r e)} & Entity set & Inner join of e with r's second elements \\
\hline
\textbf{(R r)} & (Entity, Entity) set & Reverses each tuple (x, y) to (y, x) \\
\hline
\textbf{(AND e1 e2 ...)} & Entity set & Intersection of input sets \\
\hline
\textbf{*(VALUES v1 v2...)} & Value set & Set containing values v1, v2, etc. \\
\hline
\textbf{*(IS\_TRUE s p o)} & Boolean & True if triple (s, p, o) exists \\
\hline
\textbf{*(OR e1 e2 ...) }& Entity set & Union of input sets \\
\hline
\textbf{(COUNT e)} & Integer & Cardinality of set e \\
\hline
\textbf{*(DISTINCT u)} & Entity set & Deduplicated version of set u \\
\hline
\textbf{*(GROUP\_COUNT u)} & (Entity, Value) set & Counts of distinct entities \\
\hline
\textbf{*(GROUP\_SUM gc1 gc2)} & (Entity, Value) set & Sums values from two group counts \\
\hline
\textbf{*(ALL b1 b2 ...)} & Boolean & True if all inputs are true \\
\hline
\textbf{(ARGMAX gc)} & Entity set & x where (x,y) has maximal y \\
\hline
\textbf{(ARGMIN gc)} & Entity set & x where (x,y) has minimal y \\
\hline
\textbf{(LT gc n)} & Entity set & x where (x,y) in gc and y \textless  n \\
\hline
\textbf{(LE gc n)} & Entity set & x where (x,y) in gc and y $\leq$ n \\
\hline
\textbf{(GT gc n)} & Entity set & x where (x,y) in gc and y \textgreater  n \\
\hline
\textbf{(GE gc n)} & Entity set & x where (x,y) in gc and y $\geq$ n \\
\hline
\textbf{*(EQ gc n)} & Entity set & x where (x,y) in gc and y = n \\
\hline
\end{tabular}} 
\medskip
\footnotesize{Note: * indicates new S-expression functions introduced in this work.}
\end{table}

In Table~\ref{tab:s-expr-functions}, we introduce a method to extract an S-expression core that captures the essential semantics of a natural language question. This decomposition simplifies the generation process by first extracting relatively independent substructures, which can then be combined by instantiating a predefined logical template (template composition).

\begin{figure}[ht]
    \centering
    \includegraphics[width=\textwidth]{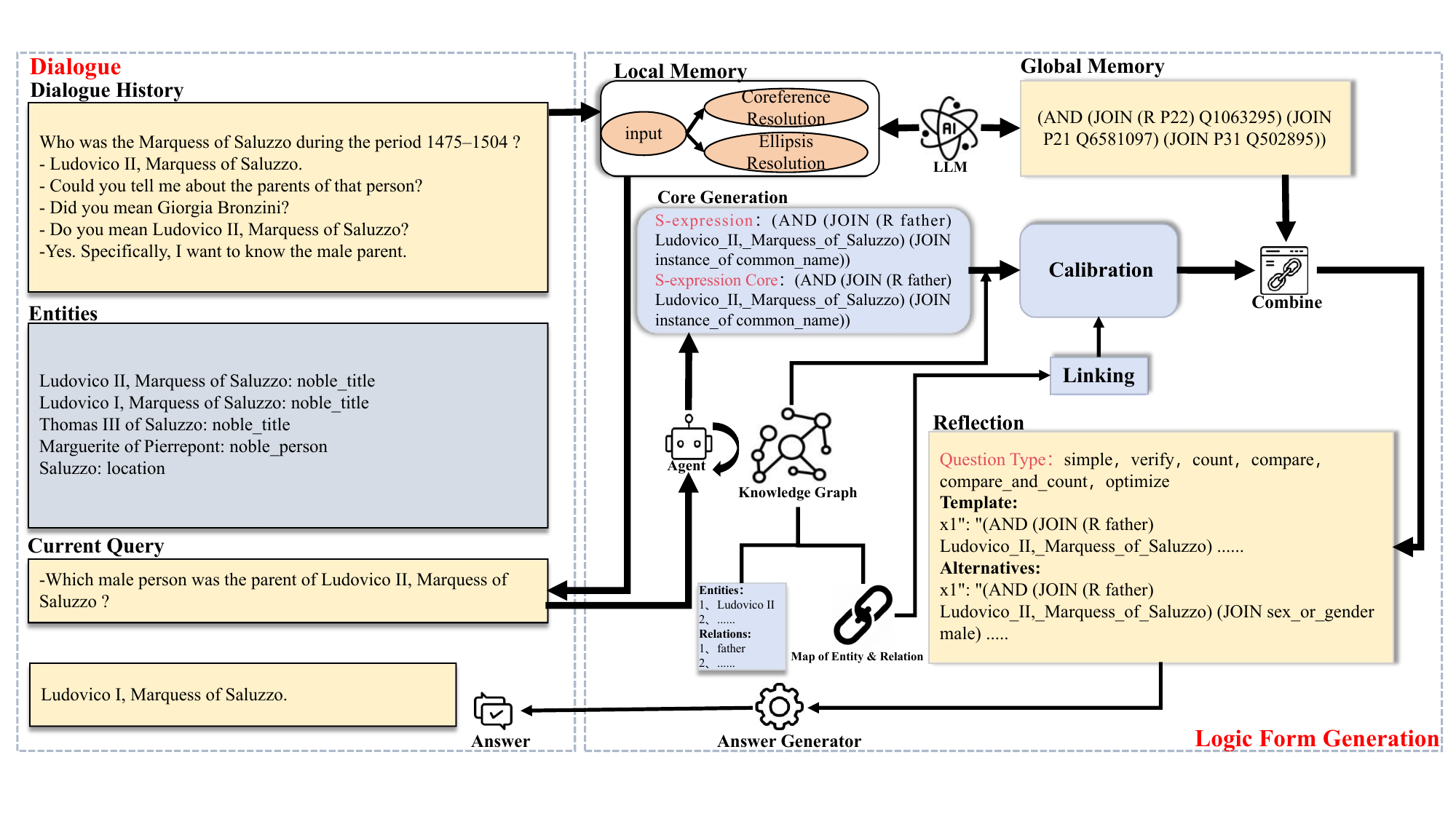}
    \captionsetup{skip=4pt}
    \caption{The framework of our method.}
    \label{figure:framework}
\end{figure}

The generation process follows two stages as shown in Figure~\ref{figure:framework}. First, the LLM generates candidate S-expression cores. An S-expression core is a simplified substructure of an S-expression which is composed of basic operations. These cores are calibrated by an agent interacting with the knowledge graph to produce refined variants. Second, the question type is predicted to select an appropriate template, and the LLM fills placeholders with functions, constants, or core expressions to produce the final S-expression.

% \begin{figure}[ht]
%     \centering
%     \includegraphics[width=1\linewidth]{img/frameworknewnew.pdf}
%     \caption{\revised{The framework of our method.}}
%     \label{figure:framework}
% \end{figure}

\subsection{Reasoning Module}

The core extraction phase, the initial critical step of the proposed method, focuses on deriving the S-expression core that encapsulates the essential semantics of natural language questions. This phase comprises two key steps:

\begin{itemize}
    \item \textbf{S-expression Core Generation}: LLM analyzes the question text to identify independent query objects, employing five fundamental functions: JOIN, R, AND, VALUES, and IS\_TRUE to articulate their logical relationships, thereby generating the S-expression core.
    \item \textbf{S-expression Core Calibration}: An agent interfacing with the knowledge graph refines the generated S-expression core by correcting syntactic errors and aligning entities and relations with the knowledge graph, yielding candidate variants.
\end{itemize}

The key innovation of this phase lies in decomposing the complex task of S-expression generation into independent substructure extractions, establishing a foundation for subsequent template integration, for details regarding specific expressions, refer to ~\ref{sec:appendixb}. Moreover, experimental validation confirms that this phased approach substantially reduces model learning complexity and enhances generation accuracy.
%template
% The template composition stage is the second key component of the proposed method. Its primary task is to combine the extracted S-expression cores into complete and executable S-expressions. This stage involves the following critical steps: question type prediction and template selection with replacement plan generation. Question type prediction is performed through intent analysis and keyword matching, categorizing questions into simple classes such as \textit{simple}, \textit{verify}, etc. Based on the predicted question type, suitable candidate templates are filtered from a predefined template library. Template selection and replacement plan generation then proceed by choosing the candidate template that best fits the question’s logic. The S-expression core is combined with the selected template, and placeholders within the template are replaced aordingly to generate the final S-expression.

% Detailing S-expression core generation
\subsubsection{S-expression Core Generation}
We introduce the concept of the S-expression core, referring to a simplified subclass of S-expression structures, the specific patterns that may appear in the core of S-expressions are shown in \ref{sec:appendixc}. Such cores typically involve only basic logical functions, such as JOIN, R, AND, VALUES, and IS\_TRUE. In the context of natural language questions, an S-expression core generally corresponds to the queried objects or targets within the question. In SPARQL queries, it maps to the graph patterns inside the WHERE\{\ldots\} clause.

%阐述s表达式核心和实体，知识图谱的一些关系以及起到的一些作用，阐述这个s表达式核心的重要性在文章中。 明确指出 core 中的实体/关系来自 KG里的E和R。
The entities and relations referenced in the S-expression core (e.g., as arguments in \texttt{JOIN} or \texttt{AND}) are elements of the underlying KG ($G = (E, R, T)$). While the LLM initially generates tokens, the subsequent calibration step aligns these with grounded entities $e \in E$ and relations $r \in R$ through KG linking, ensuring semantic consistency with the structured knowledge base.

To formally describe this generation process, we denote a natural language question as $q$ in Equation~\ref{eq:coregene1}. Through a constructed prompt $P_q = \text{Prompt}(q)$, the input is provided to a LLM. Under the parameter space $\Theta_{\text{LLM}}$, the LLM generates an S-expression core sequence $\text{Core}^* = (s_1, s_2, \ldots, s_L)$ by maximizing the conditional likelihood:

\begin{equation}
\text{Core}^* = \arg \max_{\text{Core}} P(\text{Core} \mid P_q; \Theta_{\text{LLM}})
\label{eq:coregene1}
\end{equation}
Where $q$ represents the original natural language question, $P_q$ is the prompt constructed from $q$ for LLM inference, $\text{Core}$ denotes a candidate sequence of S-expression tokens, and $\text{Core}^*$ is the optimal core sequence selected by the model. The parameter $\Theta_{\text{LLM}}$ represents all learnable parameters of the LLM, optimized during pretraining and fine-tuning to capture statistical patterns in natural language. Each $s_l$ in the sequence $(s_1, s_2, \ldots, s_L)$ corresponds to a token in the S-expression core, such as a function, an entity, a relation, or a constant. $L$ is the length of the sequence.

The output of this generation process, denoted as $\text{Core}^*$, is a preliminary representation of the question's intent. However, due to the inherent ambiguity of natural language and potential hallucinations in LLM outputs, this raw core may contain syntactic errors, unlinked surface forms, or incorrect function compositions. To address these issues, we decompose the overall core generation into two key phases. First, the LLM analyzes the question to identify independent query objects and synthesizes a preliminary core expression. Second, an agent refines this candidate by calibrating the syntactic structure and aligning its entities and relations with the underlying knowledge graph, producing multiple valid variants.

\subsubsection{S-expression Core Calibration}
\subsubsubsection{Light Linking}

Targeting KBCQA tasks~\cite{perez-beltrachiniSemanticParsingConversational2023a}, we implement conversions between extended S-expressions and SPARQL, enabling the transformation of SPARQL queries into corresponding S-expressions in the context of  KBCQA tasks. Testing confirms that all SPARQL queries in the dataset can be successfully converted to S-expressions, which can then be converted back to SPARQL while maintaining consistency with the original query results.

The proposed method employs a LLM to generate an initial S-expression draft based on the input question and annotated examples. As the LLM lacks access to the underlying knowledge graph, it produces element representations using surface names rather than canonical entity or relation identifiers. Consequently, the draft cannot be directly executed as a SPARQL query and requires a subsequent linking process. A lightweight linking strategy is adopted, which first maps each surface-named element to the most semantically similar entity or relation in the knowledge graph. Corrections are then applied to address two common error types which  are relationship inversion and type constraint errors. After completing the linking, the final S-expression is obtained and translated into an executable SPARQL query via a custom conversion function, enabling retrieval of the final answer from the knowledge graph.

In the entity and relation linking phase, this method discards the traditional entity candidate approach, retaining only the single candidate with the highest semantic similarity. Specifically, the embedding model encodes knowledge graph elements into vectors, retrieving the best match via cosine similarity. Given that LLM-generated S-expressions are semantically precise, single entity candidate suffices without degrading linking performance. 
In contrast, the entity candidate approach which keeps three entities can produce non-empty but semantically incorrect queries, masking linking errors and leading to wrong answers. The single entity candidate strategy, by comparison, enforces stricter semantic alignment and thus ensures higher consistency.

\subsubsubsection{Main Procedure of Core-Calibaration}

After the LLM generates an S-expression core $\text{Core}_j$, a calibration phase ensures both syntactic correctness and semantic grounding within the knowledge graph. This process consists of two key steps.

The first step performs syntactic correction. The agent parses the initially generated S-expression core $\text{Core}_j$ and corrects any syntactic errors to produce a structurally valid variant $\text{Core}'_j$. The correction function $\text{Corr}_{\text{syn}}(\cdot)$ detects mismatched parentheses, function argument errors, or illegal nesting structures, returning a syntactically well-formed expression:
\begin{equation}
\text{Core}'_j = \text{Corr}_{\text{syn}}(\text{Core}_j)
\label{eq:corecali1}
\end{equation}
In Equation~\ref{eq:corecali1}, $\text{Core}_j$ denotes the initial S-expression core generated by the LLM, and $\text{Core}'_j$ represents the corrected output. The correction function $\text{Corr}_{\text{syn}}(\cdot)$ performs rule-based structural validation to fix mismatched parentheses, incorrect function arity, or illegal nesting patterns.

The second step conducts knowledge graph linking to replace each surface name $s$ within $\text{Core}'_j$, such as entity or relation names, with the most semantically similar element $x^*$ selected from the candidate set $C_s$, which is retrieved from the knowledge graph ${G} = (E, R, {F})$. The optimal match is determined by computing cosine similarity over vector embeddings in Equation~\ref{eq:corecali2}:
\begin{equation}
x^* = \arg\max_{x \in C_s} \text{Cos}(\text{Embed}(s), \text{Embed}(x))
\label{eq:corecali2}
\end{equation}
The embedding function $\text{Embed}(\cdot)$ converts surface names or knowledge graph elements into vector representations, typically implemented via pretrained language models such as an  embedding model. The similarity function $\text{Cos}(\cdot, \cdot)$ measures alignment between vectors, with higher values indicating stronger semantic similarity. The candidate set $C_s$ consists of entities or relations from ${G}$ potentially corresponding to the surface form $s$.
Each $x^*$ is the best-matching element from $C_s$ according to embedding similarity.

Each surface element $s_m$ in $\text{Core}'_j$ is substituted with its optimal match $x^*_m$ to form calibrated candidates. The final set ${Calibrated\_Core}_j$ retains the candidate variants whose query executions return non-empty results. In this study, the candidate variants are preserved.

% This calibration strategy enhances alignment between LLM-generated symbolic structures and the underlying knowledge graph. Compared to standard lightweight linking methods, this approach offers two key improvements: (1) Candidate matching enhances robustness in ambiguous cases, (2) Query-executability constraints ensure semantic validity of the resulting variants.
This calibration strategy enhances alignment between LLM-generated symbolic structures and the underlying knowledge graph by building directly upon the core-wise linking mechanism described earlier. Rather than treating the entire S-expression as a monolithic unit, calibration operates on decomposed cores, leveraging the linking strategy to ground each component’s entities and relations into the knowledge graph. This design not only ensures syntactic well-formedness but also enforces semantic validity through query executability: only those variants that yield non-empty SPARQL results are retained as plausible candidates. By integrating linking as a foundational step within calibration, SEAL achieves a tighter coupling between symbolic reasoning and knowledge graph interaction, enabling robust and scalable semantic parsing in conversational settings.

\subsection{Memory Module}
We divide memory into local and global components to capture different types of information in dialog understanding. The local memory focuses on short-term contextual dependencies within the current conversation, such as coreference resolution and intent tracking. In contrast, the global memory stores structured knowledge accumulated over past interactions, enabling generalization and long-term reasoning.
\subsubsection{Local Memory}
For KBCQA tasks, coreference and ellipsis phenomena present key challenges. To ensure accurate interpretation of user intent by the LLM, our method employs the LLM for coreference resolution. Specifically, historical dialog records are provided to the LLM, enabling it to complete the user's latest question into a fully specified form based on contextual information. Specific examples of input and output can be found in \ref{sec:appendixA}. While simply concatenating historical dialogs can also complete  semantics, redundant information may degrade the accuracy of keyword matching in the subsequent question type prediction phase. Moreover, regardless of the approach, the LLM is ultimately required to resolve coreferences and ellipses. Therefore, performing semantic completion using the LLM in advance shifts critical parsing steps earlier in the pipeline, thereby enhancing overall processing efficiency.

\subsubsection{Global Memory}
The template composition stage serves as a pivotal component of the proposed framework, aiming to synthesize calibrated S-expression cores with predefined templates to construct complete and executable S-expressions. This stage comprises three primary steps: question type prediction, template selection, and replacement plan generation.

Question type prediction employs intent analysis and keyword matching to categorize questions into predefined types such as ``simple" or ``verify" This categorization constrains the search space to appropriate templates retrieved from a curated template library, which is organized according to common reasoning patterns. All generated templates are listed in \ref{sec:appendixd}.

Template selection further refines the candidate space by analyzing the logical structure of the question and selecting a suitable template $Template^*$ compatible with the calibrated S-expression core. Following this selection, a replacement plan is constructed to determine the correspondence between placeholders and concrete values. The replacement plan is formally defined as:
\begin{equation}
Plan^* = \{(P_j, Value_j) \mid j = 1, \ldots, M\}
\label{eq:localmemory1}
\end{equation}
In Equation~\ref{eq:localmemory1}, $P_j$ denotes placeholders in the template, while $Value_j$ represents the substitution elements drawn from $\{{Constants}\} \cup \{{Functions}\} \cup \{{Calibrated\_Core}_k\}$. $M$ is the number of placeholder-value pairs in the replacement plan. These elements consist of calibrated cores, functions, and constants.

The generation of the final S-expression is achieved by applying the replacement plan to the selected template, as expressed by:
\begin{equation}
S\text{-}expression_{\text{final}} = Transform(Template^*, Plan^*)
\label{eq:localmemory2}
\end{equation}
In Equation~\ref{eq:localmemory2}, $Transform(\cdot)$ is a recursive function that replaces each placeholder in $Template^*$ according to $Plan^*$ to produce the final S-expression.

The recursive transformation procedure can be further formalized as:
\begin{equation}
Transform(N) = F(Transform(N_1), \ldots, Transform(N_k))
\label{eq:localmemory3}
\end{equation}
Where the function $F$ corresponds to the operation associated with node $N$ within the template, and $N_1, \ldots, N_k$ denote its child nodes, each recursively transformed based on the plan in Equation~\ref{eq:localmemory3}.

The design of S-expression templates is grounded in a detailed analysis of the training corpus. Expression cores, specific functions , and constants are abstracted into placeholders , resulting in a versatile and reusable template library. These templates encapsulate typical logical operations such as set union, difference, deduplication, grouping, counting, and extremum computation, effectively reducing syntactic and semantic errors in LLM-generated outputs.

Adopting a template-based strategy facilitates generalization and reduces reliance on exhaustive learning by encouraging consistency through reusable syntactic patterns. In KBCQA tasks, certain equality comparison queries in SPARQL references often omit necessary FILTER components, resulting in structurally incomplete S-expressions. This observation highlights the importance of ensuring query completeness during template design to enhance the robustness of generated expressions in practical scenarios.

Crucially, in SEAL, these predefined templates serve as initial priors (or soft constraints) rather than rigid, hard-coded rules. They are designed to scaffold the reasoning process by reducing the search space for common query structures and ensuring high syntactic validity. However, the ``agent'' retains the flexibility to perform planning beyond this set. When semantic complexity demands it, the LLM can generate S-expressions that deviate from these patterns, effectively treating templates as a guiding framework rather than an absolute limit.

\subsection{Reflection Module}

Each S-expression template is associated with a predefined question type, allowing efficient type-based filtering during inference. Question type prediction is performed by prompting the LLM with three examples per type. Although the task appears straightforward, semantic overlap frequently causes confusion, particularly among types involving numerical reasoning.
To address this issue, a hybrid strategy is employed that integrates keyword-based heuristics to refine ambiguous type predictions—for example, reclassifying certain comparison-oriented questions as count-augmented variants when numerical quantification cues are detected in the utterance.

Given the predicted type, the optimal template and its corresponding replacement plan are determined by maximizing the conditional probability over the candidate space :

\begin{equation}
(\text{Tem}^*, \text{Plan}^*) = 
\arg\max_{(T, P) \in T_{\text{c}} \times P(T)} 
P\Big((T, P) \mid Q_{\text{com}}, \{\text{C\_Core}_k\}, \text{Exa};\ \Theta^{\text{ref}}_{\text{LLM}}\Big)
\label{eq:reflection1}
\end{equation}
In Equation~\ref{eq:reflection1}, the candidate template set $T_{\text{c}}$, comprising templates $\text{Tem}_m$, is determined through filtering based on the predicted question type. For a given template $T$, $P(T)$ denotes the set of all possible replacement plans compatible with $T$. Each $P \in P(T)$ represents a specific assignment of values to the placeholders in $T$, and the pair $(T, P)$ denotes a candidate template-plan combination considered during selection. $Q_{\text{com}}$ is the complete form of the current question, $\text{C\_Core}_k$ denotes the set of calibrated S-expression cores from the current dialog, $\text{Exa}$ represents the set of in-context examples, and $\Theta^{\text{ref}}_{\text{LLM}}$ denotes the parameters of the LLM used in the reflection phase. Examples of related inputs and outputs can be found in \ref{sec:appendixf}.

To resolve potential ambiguity among compare\_and\_count, compare, and count, heuristic-based adjustment rules are introduced. In Equation~\ref{eq:reflection2}, $\text{Type}_{\text{final}}$ is the corrected question type after heuristic adjustment, ${H}(\cdot)$ is a rule-based function that refines the initial prediction by analyzing keyword patterns and syntactic cues in $Q_{\text{com}}$, $\text{Type}_{\text{pre}}$ denotes the initially predicted type, and $Q_{\text{com}}$ is the complete natural language form of the current question:

\begin{equation}
\text{Type}_{\text{final}} = {H}(\text{Type}_{\text{pre}}, Q_{\text{com}})
\label{eq:reflection2}
\end{equation}
The LLM is provided with up to four examples per candidate template to facilitate accurate template selection. This structured methodology simplifies the selection process while leveraging the semantic transparency of S-expressions.

Formally, we define the high confidence criterion for memory updates as a composite state satisfying three conditions: (1) Maximal Linking Score, where entities/relations achieve the highest cosine similarity in dense retrieval; (2) Execution Validity, requiring non-empty SPARQL results; and (3) Semantic Alignment, where the reflection module confirms that the logical predicates (e.g., ``instance\_of") do not violate implicit constraints in the natural language query. Only queries satisfying all three are serialized into the global Memory.

\subsection{Self-Evolving Mechanism}

\begin{figure}[ht]
    \centering
    \includegraphics[width=1\linewidth]{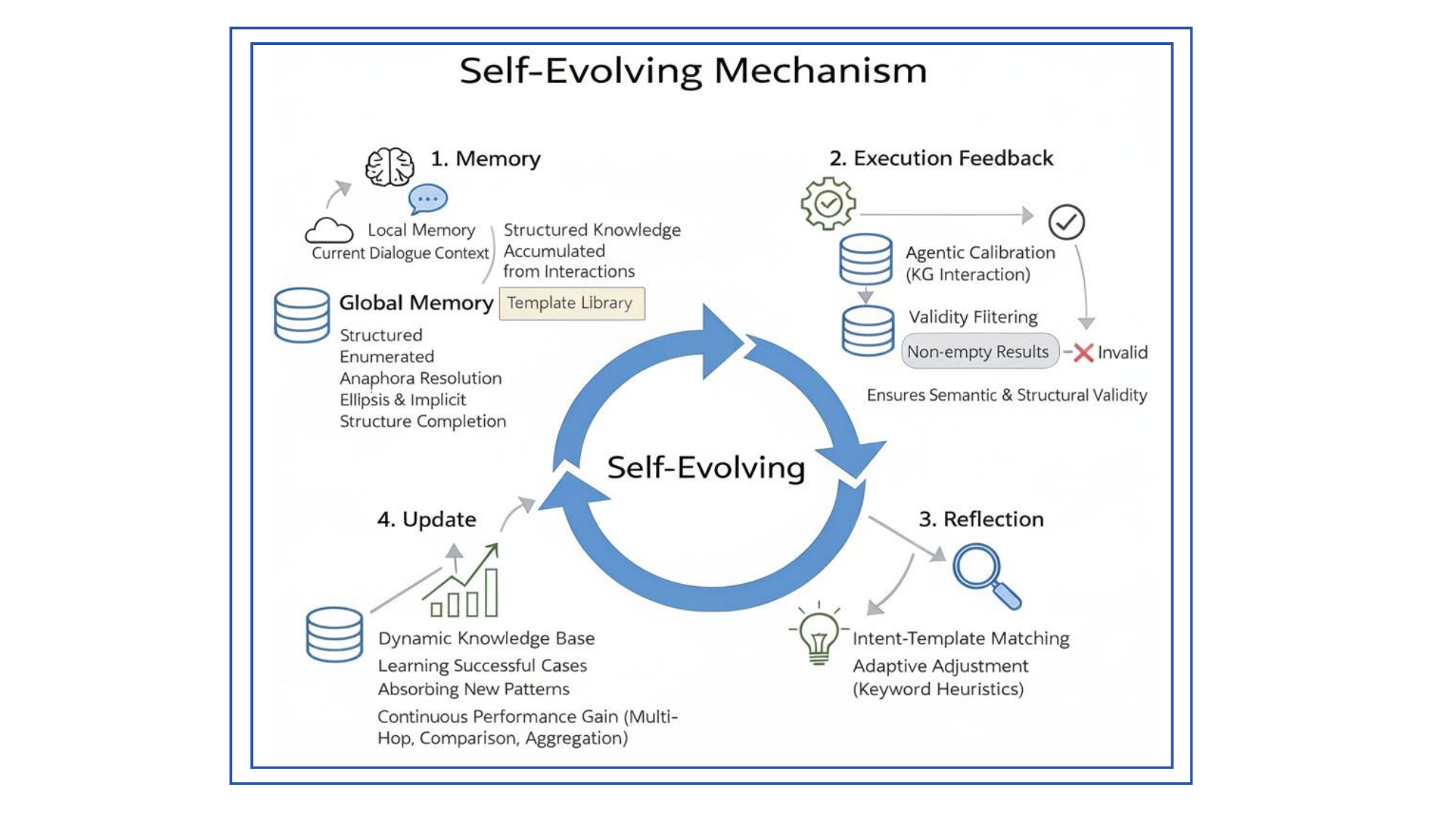}
    \caption{Self-Evolving Mechanism}
    \label{figure:self_evolving}
\end{figure}

Most existing methods rely on fixed KG and static parsing rules. Such systems struggle to effectively adapt to the novel expressions continuously emerging in real-world dialog. To overcome this limitation, we introduce a self-evolving mechanism. It establishes a continuously learning agent through the close synergy of local memory, global memory, and a reflection module. In each dialog turn, the local memory module maintains the current context state, which includes resolved entities and completed semantic intent. This maintenance ensures precise input comprehension and semantic consistency, particularly when handling multi-turn dependencies. The global memory module structurally stores knowledge distilled from successful past dialogs. Knowledge is typically organized as question type to relevant question sample pairs. This mechanism supports the long-term reuse of historically validated logical forms for frequent queries. It guides S-expression generation, significantly boosting parsing accuracy and query efficiency. Following S-expression execution, the reflection module performs post-analysis. Analysis includes syntactic validation, predicate-relation alignment checks, and result non-nullity tests. If this module detects errors, such as entity linking failure or structural invalidity, it logs the cause. It then triggers a correction loop to generate an alternative logical form in Figure \ref{figure:self_evolving}.

The reflection module validates a generated S-expression via LLM-as-Judge manner~\cite{zhong2025comprehensive} to check if it is syntactically well-formed, exhibits consistent alignment between natural language predicates and knowledge graph relations, and executes to produce a non-empty result. Upon such validation, the S-expression, along with its question type and surface-form pattern, is serialized and incorporated into the global memory as a new template instance. 
To ensure scalability and prevent memory degradation over long-term interactions, the Global Memory stores abstracted templates (with variables replacing specific entities) rather than raw queries. This abstraction naturally limits memory growth, as the number of valid logical structures is significantly smaller than the number of unique questions. Furthermore, a frequency-based deduplication strategy is employed, where redundant patterns are merged, and low-frequency error-prone templates are implicitly pruned over time.

This update is performed incrementally and selectively: only execution-verified logical forms with high confidence are retained, ensuring that the global memory evolves through the accumulation of reliable and reusable knowledge rather than unverified or noisy hypotheses.

Crucially, this mechanism allows the system to capture novel, successful out-of-template structures that may be generated during the calibration phase (e.g., nested logic not present in the initial library). By verifying and storing these novel structures, the Global Memory effectively expands the template library dynamically. This transforms the system from a static template-filler into an evolving agent that learns the grammar of S-expressions over time.

The self-evolving mechanism transforms global memory from static storage into a dynamically updateable KG. Validated knowledge is incrementally written through reflection. 
This closed-loop process, which involves perceiving the current input, retrieving relevant historical patterns, reflecting on execution outcomes, and updating the KG, enables adaptive learning without explicit retraining. 
As dialog progresses, the system’s capacity to handle similar or complex queries progressively strengthens. This demonstrates sustained evolutionary performance.

First, SEAL's self-evolving mechanism fundamentally represents a structured knowledge update based on logical verification. Unlike traditional RAG strategies that match similar text at a semantic level, SEAL's global memory exclusively stores high-confidence logic forms (S-expressions) verified through execution on the KG. This means the system corrects failed reasoning paths via the 'reflection module' in every interaction and converts successful reasoning paradigms into structured knowledge.
Second, SEAL provides a closed-loop iterative parsing strategy. The calibration module is not merely a heuristic correction tool but acts as a dynamic planner combining structured experience from global memory. This ``Generate-Execute-Evaluate-Reflect-Store" closed loop enables the agent to cope with long-tail and complex logical combinations by accumulating experience without parameter updates, which is closer to human experiential learning than simple engineering retrieval.

\begin{algorithm}[ht!]
\caption{Self-Evolving Process}
\label{alg:seal_process}
\begin{algorithmic}[1]
\Require 
    Current question $q_i$;
    
    Dialogue history $H_{i-1} = (q_1, a_1, \dots, q_{i-1}, a_{i-1})$; 
    
    Knowledge Graph $G = (E, R, T)$; 
    
    Global Memory $M_{global}$.
\Ensure 
    Answer $a_i$; 
    Updated Global Memory $M_{global}'$.

\State \textbf{Initialize:} $S_{final} \leftarrow \text{null}$, $Result \leftarrow \text{null}$

% \Statex \Comment{\textbf{\textit{Stage 1: Local Memory \& Context Perception}}}
\State $q_{com} \leftarrow \text{CoreferenceResolution}(q_i, H_{i-1}, \Theta_{LLM})$ %\Comment{Resolve implicit references}
\State $State_{local} \leftarrow \{ResolvedEntities, Intent\}$% \Comment{Update local context state}

% \Statex \Comment{\textbf{\textit{Stage 2: Retrieval from Global Memory}}}
\State $Type_{pre} \leftarrow \text{PredictType}(q_{com})$
\State $(T_c, Exa) \leftarrow \text{Retrieve}(M_{global}, Type_{pre})$ %\Comment{Retrieve templates \& examples}

% \Statex \Comment{\textbf{\textit{Stage 3: Reasoning \& Calibration}}}
\State $Core_{raw} \leftarrow \text{GenerateCore}(q_{com}, \Theta_{LLM})$ %\Comment{LLM generates S-expr core}
\State $Core_{calib} \leftarrow \text{Calibrate}(Core_{raw}, G)$ %\Comment{Agent corrects syntax \& aligns with KG}
\State $(Tem^*, Plan^*) \leftarrow \arg \max_{(T,P)} P((T, P) \mid q_{com}, Core_{calib}, Exa)$ %\Comment{Eq. 8}
\State $S_{final} \leftarrow \text{Transform}(Tem^*, Plan^*)$ %\Comment{Eq. 6}

% \Statex \Comment{\textbf{\textit{Stage 4: Execution \& Reflection Loop}}}
\While{$RetryCount < MaxRetries$ \textbf{and} $Result$ is null}
    \State $Result \leftarrow \text{ExecuteSPARQL}(S_{final}, G)$
    \State $Check_{syn} \leftarrow \text{ValidateSyntax}(S_{final})$
    \State $Check_{align} \leftarrow \text{ValidateAlignment}(S_{final}, q_{com})$
    \State $Check_{non\_empty} \leftarrow (Result \neq \emptyset)$
    
    \If{$Check_{syn} \land Check_{align} \land Check_{non\_empty}$}
        \State \textbf{break} %\Comment{Validation passed}
    \Else
        \State $ErrorLog \leftarrow \text{Refine}(Result, S_{final})$ %logerror改一下
        \State $S_{final} \leftarrow \text{CorrectionLoop}(S_{final}, Refine, G)$ %\Comment{Self-correction}
    \EndIf
\EndWhile

% \Statex \Comment{\textbf{\textit{Stage 5: Self-Evolution (Global Memory Update)}}}
\If{$Result \neq \emptyset$ \textbf{and} $Check_{valid}$ is \textbf{true}}
    \State $Pattern_{new} \leftarrow \text{ExtractSurfacePattern}(q_{com})$
    \State $Entry_{new} \leftarrow (Type_{pre}, Pattern_{new}, S_{final})$
    \State $M_{global}' \leftarrow M_{global} \cup \{Entry_{new}\}$ %\Comment{Incrementally update knowledge}
\EndIf

\State \Return $Result, M_{global}'$
\end{algorithmic}
\end{algorithm}

\section{Experiments} % DONE
\label{sec:exp}
\subsection{Experimental Setup}
In this section, we first introduce the datasets and evaluation metrics used for evaluation. Then, we present baseline methods for comparison and finally explain the implementation details.

\textbf{Datasets.} 
We conduct the experiments on SPICE\cite{perez-beltrachiniSemanticParsingConversational2023a}, a conversational semantic parsing dataset over Wikidata~\cite{DBLP:conf/www/Vrandecic12} derived from CSQA~\cite{ComplexSequentialQuestion} benchmark. 
% Conversation-level
Each conversation instance in the SPICE dataset is a natural language user-system QA sequence.
% Question-level
Additionally, SPARQL parsing for mapping natural language to KG query statements is also provided, exploring the paths and entities on the underlying KG.
% Our partition
For experiments, we select 9 of the 10 types of questions from SPICE, covering simple questions, logical reasoning, and comparative reasoning. We discard the ``Clarification" subset as it lacks corresponding SPARQL queries. 
To reduce cost, we randomly sample conversation instances from the complete conversation, meanwhile ensuring that each subtype of questions contains at least 50 samples. 

\textbf{Evaluation Metrics.} % DONE
% Brief Intro 
Following previous works~\cite{perez-beltrachiniSemanticParsingConversational2023a}, we choose marco-F1 and Accuracy score as the execution-based evaluation metrics for the experiments, which are used to evaluate the questions with answer types of entity sets and numerical values (boolean values), respectively. The marco-F1 metric averages the F1 score of each question subset to prevent the evaluation bias caused by imbalanced instance amount.

\textbf{Baselines.} % DONE
% Beief Intro
We compare SEAL with two types of baselines: 
% Supervised
For supervised methods, we utilize BertSP~\cite{perez-beltrachiniSemanticParsingConversational2023a} and DCG~\cite{DBLP:conf/emnlp/JainL23} for comparison. These approaches use the AllenNLP tool~\footnote{https://github.com/allenai/allennlp} for NER and global look-up (denoted as ${GL}$) for type linking.
% Unsupervised 
For the unsupervised type, we adapt KB-Binder~\cite{liFewshotIncontextLearning2023}, a strong single-turn semantic parsing-based KBCQA baseline to the conversational question answering setting.
Note that we select KB-Binder as our primary LLM-based semantic parsing baseline rather than other methods(e.g. KB-Coder) due to fundamental differences in task formulation (single-turn vs. multi-turn KBQA) and knowledge base structure compatibility with the SPICE dataset.
Additionally, we follow \cite{DBLP:conf/icaart/SchneiderKJSM24} to utilize LLMs to generate logic forms directly by providing them linked entities and relations.
The details of the baselines can be found in Sec~\ref{sec:rw}.

\textbf{Implementation Details.} 
The question taxonomy consists of six types: \textit{simple}, \textit{verify}, \textit{count}, \textit{compare}, \textit{compare\_and\_count}, and \textit{optimize}, following the annotation schema of the SPICE dataset. For question type prediction, we employ the Qwen2.5-32B-Instruct model prompted with three in-context examples per type. All other components—coreference resolution, S-expression core generation, template selection, and replacement plan generation—are implemented using the DeepSeek-V3 model.

Two key parameters are explored in the experiment. 
The first concerns the linking strategy during the binding of surface forms to knowledge graph elements: we compare a top-1 approach (retaining only the highest-similarity candidate) against a top-k strategy (retaining k=3 entity or relation candidates per mention, as commonly used in prior work).
The second parameter is the number of candidate variants preserved during calibration (either the first variant yielding a non-empty query result or the top three variants). 
These retention strategies influence both the accuracy and efficiency of the linking process.

\subsection{Main Results}
\subsubsection{S-expression Core Extraction-based Method}
When retaining only one variant, a comparison between selecting the single best candidate and selecting multiple candidates shows that the multiple candidate strategy issues slightly more SPARQL queries, though the difference is minimal. This result supports the earlier observation that S-expression cores generated by LLMs are generally semantically accurate.

We present the main experimental results in Table~\ref{table:main}, comparing our method \textsc{SEAL} with state-of-the-art supervised and unsupervised approaches across various question types. The bold values indicate the best performance for each task category. As shown, \textsc{SEAL} achieves competitive results in both supervised and unsupervised settings, particularly excelling in complex reasoning tasks. Specifically, in \textit{Logical Reasoning}, \textsc{SEAL} obtains a m-F1 of 73.08, surpassing all baselines and demonstrating strong capability in handling multi-hop logic. In \textit{Quantitative Reasoning}, it achieves 64.45, outperforming KB-Binder by a large margin despite lacking supervision. For \textit{Comparative Reasoning}, \textsc{SEAL} reaches 41.06, significantly higher than KB-Binder's 12.17, indicating its robustness in comparative queries. On simple questions, \textsc{SEAL} performs well across variants, achieving 78.49 on direct questions and 70.03 on ellipsis-based ones, showing effective coreference resolution. In the verification tasks, \textsc{SEAL} also achieves the highest accuracy at 85.97 for boolean verification and 70.12 for count-based comparison, further validating its generalization ability. Overall, \textsc{SEAL} achieves an AC of 66.83, significantly outperforming KB-Binder (36.66) and approaching the performance of supervised models like LLM$_{GT}$ (65.65), while requiring no labeled data.

%本论文主试验表格
\begin{table}[]\small
\caption{Main results}
\label{table:main}
\resizebox{\linewidth}{!}{
\begin{tabular}{lcccccc}
\toprule[1pt]
\multirow{2}{*}{Question Type} & \multirow{2}{*}{Case num} & \multicolumn{2}{c}{Supervised}                                    & \multicolumn{3}{c}{Unsupervised}      \\ \cline{3-7} 
              &          & \multicolumn{1}{l}{DCG$_{GL}$} & \multicolumn{1}{l}{BertSP$_{GL}$} & \multicolumn{1}{l}{LLM$_{GT}$} & \multicolumn{1}{l}{KB-Binder} & \multicolumn{1}{l}{SEAL} \\ \hline
              & {}   & {m-F1} & {m-F1}  & {m-F1}  & {m-F1}   & m-F1    \\ \hline
Logical Reasoning            & {421}  & {54.21} & {24.08}  & {89.61}  & {46.85}  & \textbf{73.08} \\
Quantitative Reasoning       & {220}  & {\textbf{89.67}} & {73.23}  & {21.01}  & {15.41}  & 64.45 \\
Comparative Reasoning        & {329}  & {\textbf{79.86}} & {71.44}  & {5.87}  & {12.17}  & 41.06 \\
Simple Question (Coref)      & {698}  & {\textbf{82.53}} & {72.19}  & {85.73}  & {41.07}  & 71.53 \\
Simple Question (Direct)     & {739}  & {\textbf{85.12}} & {68.87}  & {92.69}  & {40.71}  & 78.49 \\
Simple Question (Ellipsis)   & {181}  & {\textbf{77.57}} & {60.33}  & {61.24}  & {39.48}  & 70.03 \\ \hline
              & {}   & {AC} & {AC} & {AC}  & {AC}     & AC    \\ \hline
Verification (Boolean)          & {385}  & {73.19} & {39.45}  & {91.89}  & {64.03}  & \textbf{85.97} \\
Quantitative Reasoning (Count) & {482}  & {62.72} & {50.24}   & {58.33}  & {39.00}  & \textbf{70.12} \\
Comparative Reasoning (Count)  & {336}  & {\textbf{63.90}} & {47.52}  & {5.06}  & {7.44}   & 22.02 \\ \hline
Overall       & {3791} & {\textbf{74.72}} & {57.33}    & {65.65}   & {36.66}    & {66.83}  \\
\bottomrule[1pt]
\end{tabular}
}
\end{table}

\subsection{Efficiency Analysis}

Under the setting of retaining three candidate variants, the total number of SPARQL queries remains low despite a slight increase. Compared to KB-BINDER, our method  mitigates the long-tail issue of high query counts, demonstrating improved efficiency. This advantage stems from two key factors: (1) decomposing the full S-expression linking task into multiple core-level subtasks reduces the base of exponential growth, and (2) core expressions are semantically simpler and easier to link, making it more likely to find non-empty results early and avoid unnecessary queries, thereby further reducing SPARQL overhead.

\begin{figure}[H]
    \centering
    \includegraphics[width=.8\linewidth]{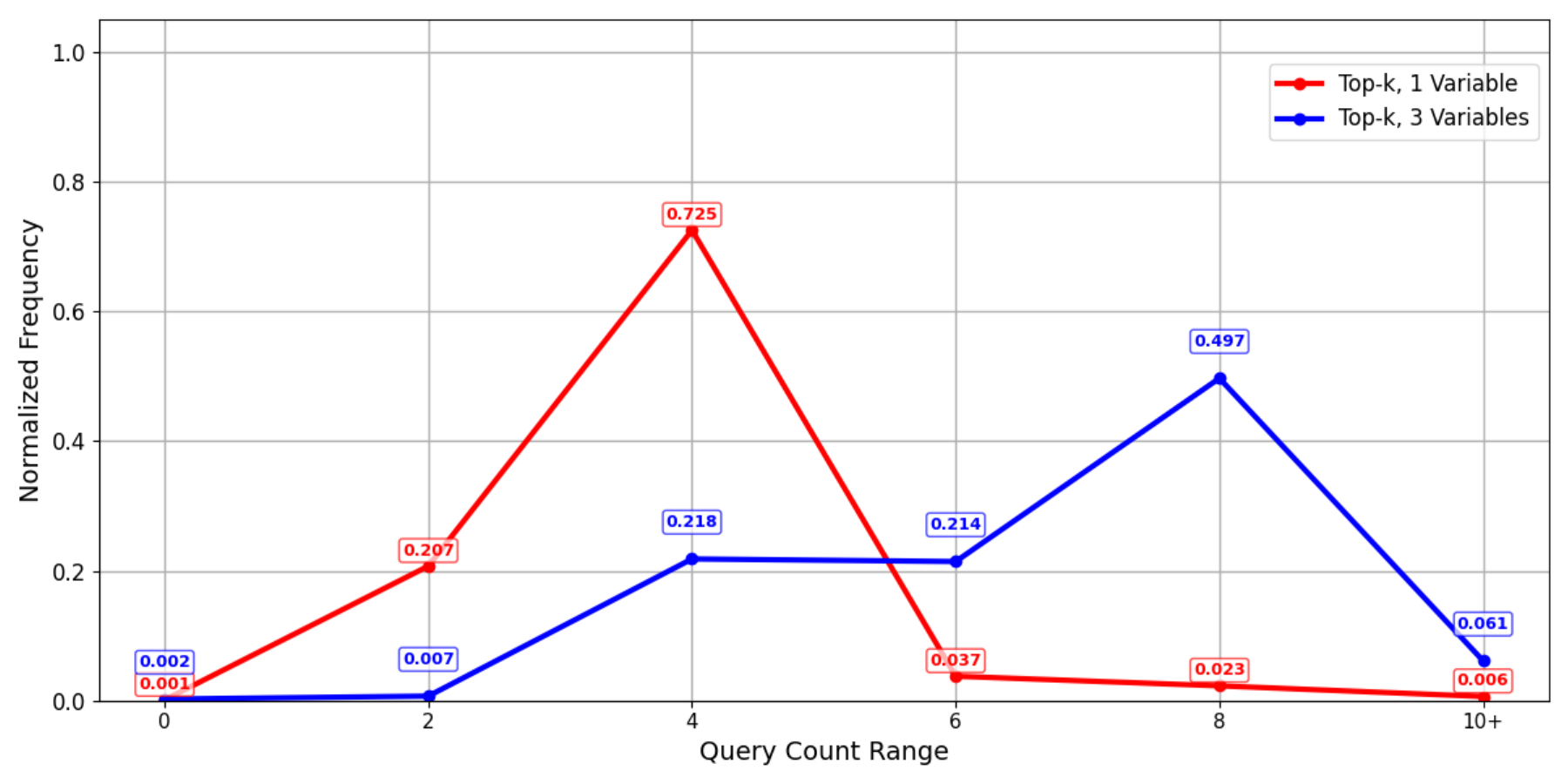}
    \caption{SPARQL query count for each S-expression core extraction parameter setting}
    \label{figure:3}
\end{figure}

As shown in Figure~\ref{figure:3}, compared to the approach of retaining a single best candidate, the method of retaining multiple candidates (k=3) improves recall by considering a broader set of options, though it slightly reduces precision due to increased noise. It performs better overall on complex questions. Building on this, retaining multiple variants further enhances recall, especially for questions involving multiple conditions or complex relations, by reducing semantic drift during the calibration phase. However, for simple questions with clear intent, retaining too many variants introduces noise, complicating the selection of the correct answer and thus lowering overall performance.

To address the computational burden associated with semantic parsing over large knowledge graphs, SEAL adds a core extraction mechanism that inherently improves inference efficiency. Theoretically, generating a complete S-expression for a complex query involves searching over a space whose size grows exponentially with the number of involved entities and relations. By decomposing the target logical form into smaller, functionally coherent sub-structures, core extraction reduces this problem to solving multiple low-dimensional linking tasks. The total search space is thus bounded by the sum rather than the product  of the individual core candidate sets, effectively transforming exponential complexity into near-linear scaling with respect to query depth.

The local memory module further enhances efficiency by resolving coreferences and completing elliptical expressions at the earliest stage of processing. This upfront disambiguation yields a self-contained, context-independent query for downstream components—such as core extraction and type prediction—eliminating the need for these modules to consider multiple contextual interpretations. Consequently, the system avoids generating and evaluating redundant candidate logical forms, reducing both computational redundancy and error propagation.

Although our framework relies on a library of predefined S-expression templates, we observe that the LLM can generalize beyond these constraints when necessary. Qualitative analysis demonstrates the agent's capability to perform planning beyond the template set. For instance, in questions requiring logical disjunction over three or more independent query cores, the model autonomously constructs nested expressions such as (OR (OR x1 x2) x3) , which are not explicitly covered by any single template in the global memory. This behavior demonstrates the LLM’s capacity for compositional synthesis and robust generalization in out-of-template scenarios. Additional examples are provided in ~\ref{sec:appendixe}.

\subsection{Structure Accuracy}
\label{sec:strucac}

As shown in Table~\ref{table:3}, the core extraction method consistently outperforms direct generation in structural accuracy. For all question types beyond Simple Questions, it achieves  higher structural overlap and parsing success rates. This is because complex S-expression structures are harder to learn directly, whereas the two-stage approach helps reduce syntax and structural errors. This leads to outputs more closely aligned with the correct expressions.

\begin{table}[htbp]
    \small
    \centering
    \caption{Comparison of core extraction and KB-BINDER (structure overlap and parsing success) (\%)}
    \renewcommand{\arraystretch}{1.2}
    \setlength{\tabcolsep}{2.5pt}
    \begin{tabular*}{\textwidth}{@{\extracolsep{\fill}}lcccc}
        \hline
        \multirow{2}{*}{Q. Type} 
 & \multicolumn{2}{c}{Struct. Overlap (\%)} 
 & \multicolumn{2}{c}{Parse Success (\%)} \\
        \cline{2-5}
 & Core Ext. & KB-BINDER 
 & Core Ext. & KB-BINDER \\
        \hline
        Simple Q. (Direct) & 71.7 & 64.7 & 97.2 & 94.9 \\
        Simple Q. (Coref.) & 67.6 & 53.3 & 96.4 & 91.8 \\
        Comp. Reasoning & 30.7 & 4.9 & 85.7 & 41.0 \\
        Comp. Reasoning (Count) & 26.2 & 0.0 & 85.7 & 24.7 \\
        Quant. Reasoning (Count) & 42.1 & 3.3 & 74.1 & 51.9 \\
        Logical Reasoning & 33.5 & 15.7 & 91.2 & 63.4 \\
        Verification & 95.6 & 26.8 & 95.8 & 48.3 \\
        Simple Q. (Ellipsis) & 61.9 & 75.1 & 97.8 & 97.8 \\
        Quant. Reasoning & 28.2 & 15.0 & 40.9 & 27.7 \\
        \hline
        Total & 54.8 & 32.2 & 88.1 & 66.0 \\
        \hline
    \end{tabular*}
    \label{table:3}
\end{table}

\subsection{Analysis of Structural Complexity and Evolutionary Adaptability}
% To measure the effectiveness of the self-evolving mechanism, we conducted a series of experiments. The experimental results are shown in Figure~\ref{fig:three-subfigures}.

\begin{figure}[htbp]
\centering
\begin{subfigure}[t]{0.325\textwidth}
  \centering
  \includegraphics[width=\linewidth]{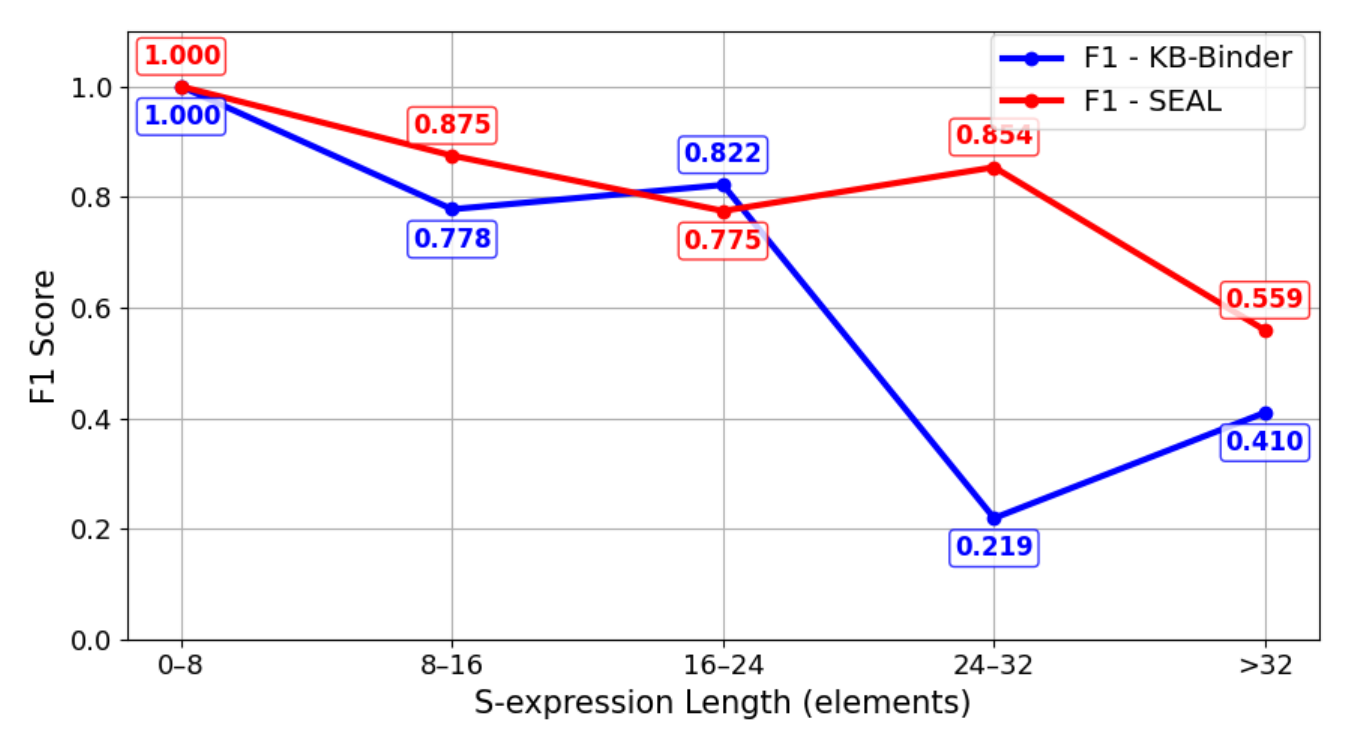}
  \caption{Impact of S-expression length}
  \label{fig:sub1}
\end{subfigure}%
\hspace{0.01\textwidth}% ← 微小间距
\begin{subfigure}[t]{0.325\textwidth}
  \centering
  \includegraphics[width=\linewidth]{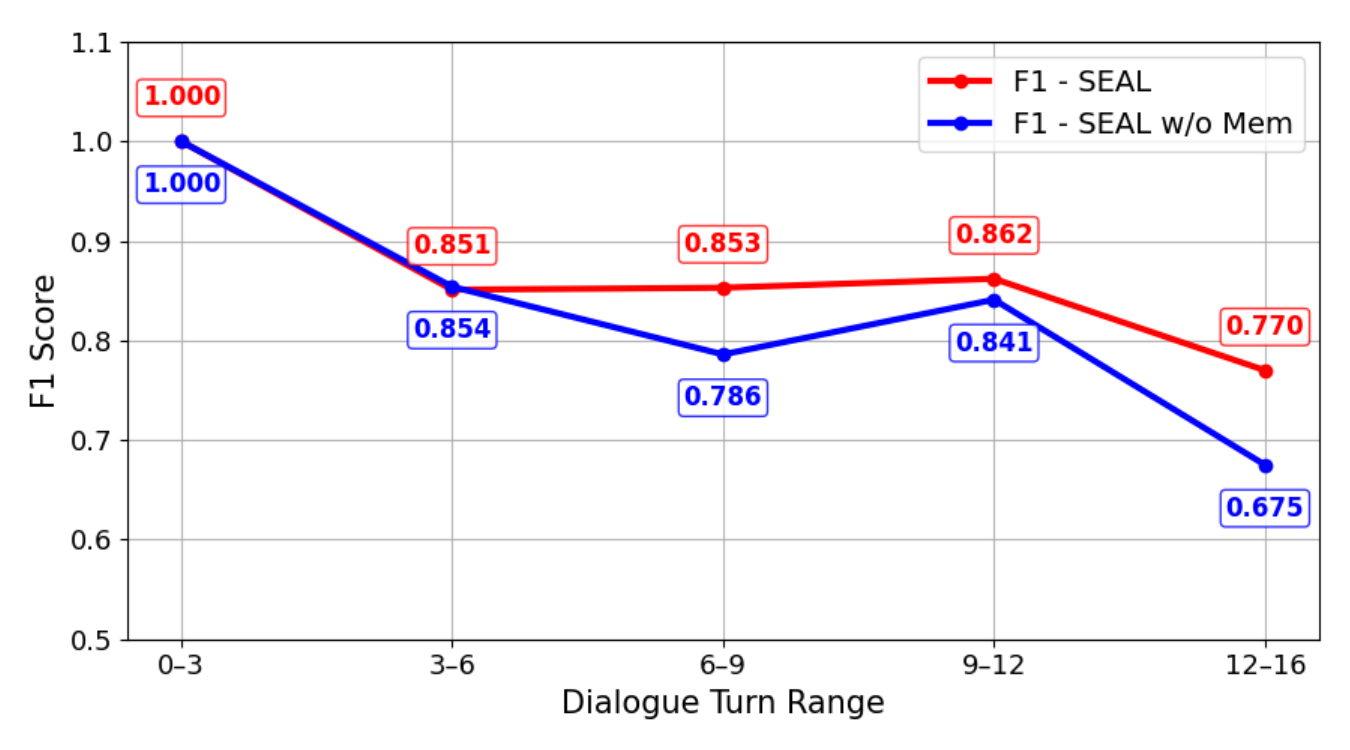}
  \caption{Impact of dialogue turn}
  \label{fig:sub2}
\end{subfigure}%
\hspace{0.01\textwidth}%
\begin{subfigure}[t]{0.325\textwidth}
  \centering
  \includegraphics[width=\linewidth]{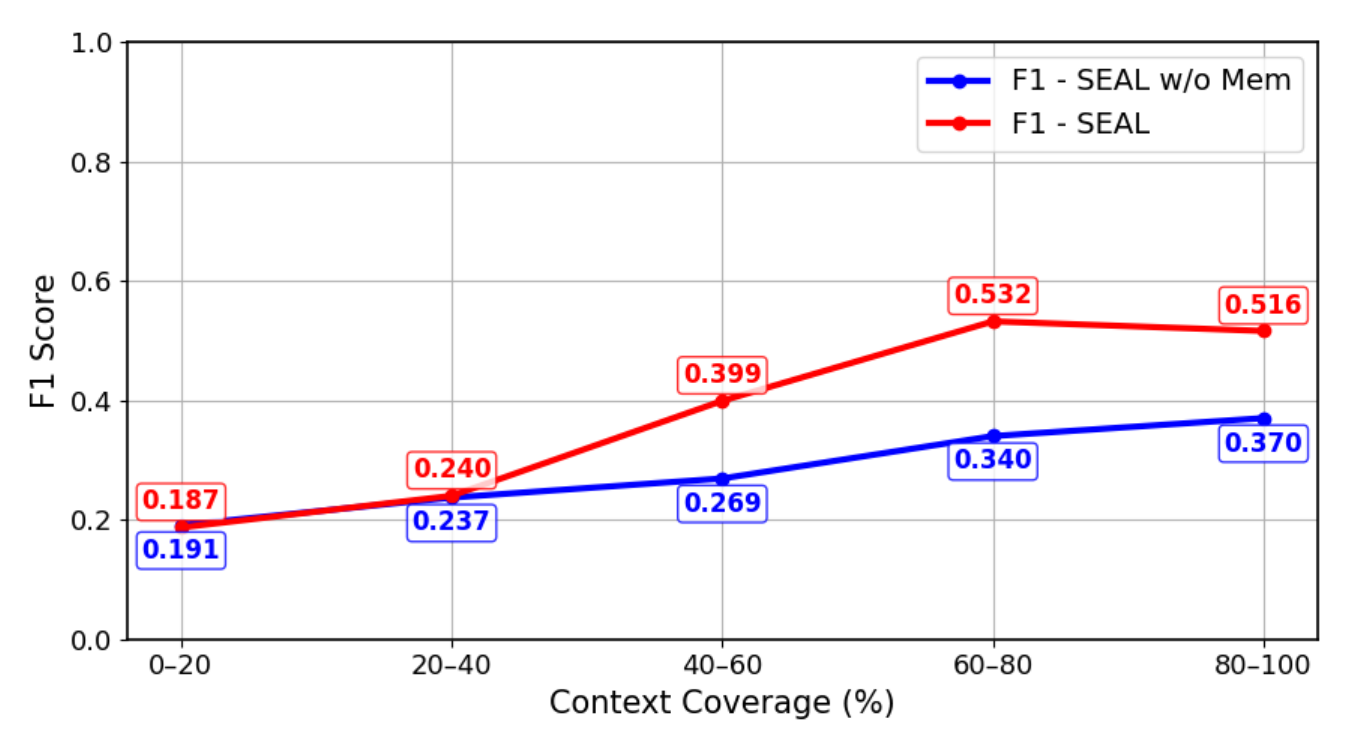}
  \caption{Impact of cumulative dialogue}
  \label{fig:sub3}
\end{subfigure}
\caption{Impact of different factors on F1 under our method. (a) S-expression length; (b) dialog turn depth; (c) cumulative dialog context coverage}
\label{fig:three-subfigures}
\end{figure}

To comprehensively evaluate the robustness of SEAL, we conducted a series of analyses focusing on two key dimensions: the static structural complexity of individual queries and the dynamic adaptability across progressing dialogue turns. The results are illustrated in Figure~\ref{fig:three-subfigures}.

First, we analyze the impact of structural complexity (Figure~\ref{fig:three-subfigures}(a)).
We segmented all test samples into intervals based on the length of S-expressions (0–8, 8–16, ..., >32) and calculated the F1 scores for SEAL and KB-Binder for each interval. The results are shown in Figure~\ref{fig:three-subfigures} (a).
The query length refers to the number of elements in the target S-expression, serving as a metric for semantic logic complexity. SEAL demonstrated superior stability and adaptability across varying complexities. While both models performed well in the short query stage (<16 elements) , KB-Binder's performance dropped sharply as complexity increased. Specifically, in the 24–32 element interval, its F1 score decreased from 0.822 to 0.219, indicating a high susceptibility to long-range dependencies and error accumulation. In contrast, F1 score of SEAL avoided this decline, maintaining 0.854 in the same range, showcasing robust parsing for complex structures. Even with extremely long expressions (>32 elements), SEAL scored  0.559, outperforming KB-Binder's 0.410 (a relative improvement of 36.3\%). 
This performance advantage on complex structures is primarily attributed to our proposed S-expression core extraction and the two-stage generation framework.
This validates the S-expression core extraction mechanism of SEAL and its modular two-stage design, which prevents the performance collapse common in end-to-end long sequence generation.

{Second, we evaluate the system's evolutionary capability in dynamic multi-turn scenarios (Figure~\ref{fig:three-subfigures}(b) and (c).

To verify whether our method possesses the ability to continually optimize performance as the dialog progresses, we segmented the entire sequence of dialog turns into several ranges (0–3, 3–6, \dots, 12–16 turns) and evaluated the F1 performance within each interval. Here, dialog turn refers to the position of the current question within the multi-turn dialog. The experimental results are shown in Figure~ \ref{fig:three-subfigures}(b). 
In the early phase (0–3 turns), the F1 of both SEAL and the ablation model (SEAL w/o memory) achieved 1.0. However, SEAL's advantage became evident from the 6th turn. In the 6–9 turn range, SEAL maintained stability (0.853), while the memory-ablated version dropped to 0.786. SEAL then peaked at 0.862 in the 9–12 turn range, contrasting sharply with the ablation model's limited rebound (0.841). In the final stage (12–16 turns), SEAL's F1 (0.770) remained superior to the ablated version's 0.675 (14\% gap). This trend confirms that SEAL achieves self-evolving behavior by using local memory for contextual consistency and global memory to retrieve successful patterns, thereby continuously enhancing semantic parsing capability during complex dialogs.

Furthermore, Figure~\ref{fig:three-subfigures}(c) illustrates the impact of cumulative context coverage.
To measure the impact of the available proportion of cumulative dialog history data (data stream) on F1 performance in multi-turn dialogs and to quantify the contribution of memory, we segmented the entire dialog dataset into several intervals based on the proportion of visible context (0–20\%, 20–40\%, 40–60\%, 60–80\%, 80–100\%). The model's F1 performance was then evaluated within each interval.
The system's performance changes based on the amount of historical dialog observed. The results show that in the early stages (<40\% data stream), the performance of both versions (with and without memory) is similar, indicating that the memory mechanism is not yet influential during the initial dialog turns. However, starting from  40\%, SEAL (with memory) shows a  performance leap, while SEAL w/o memory exhibits gradual growth. In the 60–80\% interval, SEAL's F1 peaks at 0.532, surpassing the memory-ablated version (0.340) by 56\%. Even in the final stage (80–100\% interval), where F1 values slightly decrease, likely due to noise accumulation from long dialogs, SEAL maintains a clear advantage.

% 上述结果表明,SEAL 的全局记忆机制能够有效利用对话历史中的结构化知识,在中后期对话轮次中显著提升语义解析性能。随着可见上下文比例的增加,SEAL 表现出明显的“自我演化”趋势——即通过累积经验不断优化推理能力,而移除记忆模块的消融模型则缺乏这种持续学习能力,性能增长趋于饱和。这为“系统性能随对话轮次提升”的实验现象提供了直接证据,验证了所提出 self-evolving 架构的有效性。

These results collectively demonstrate that SEAL's global memory effectively utilizes structured knowledge from the dialog history, significantly improving semantic parsing in mid-to-late turns. As visible context increases, the results show a clear benefit from self-evolving functionality as SEAL continuously improves its inference capability through accumulated experience. Conversely, lacking the memory module, the ablated model cannot sustain this learning, reflected in slower performance growth. This directly validates the proposed self-evolving architecture, showing system performance improves with increasing dialog turns.

\subsection{Ablation Study}

\begin{table}[]
\centering
\caption{Performance metrics of ablated variants}
\label{table:42}
\begin{tabular}{lccc}
\toprule[1pt]
                    & F1    & AC    & Overall \\ \hline
SEAL                  & 66.44 & 59.37 & 64.08   \\
w/o core extraction   & 41.34 & 35.39 & 39.36   \\
w/o entity candidate  & 39.32 & 46.16 & 41.60   \\
w/o calibration       & 61.78 & 56.21 & 59.93   \\
w/o local memory      & 61.89 & 56.69 & 60.16   \\
\bottomrule[1pt]
\end{tabular}
\end{table}

Table~\ref{table:42} presents the results of an ablation study on the SEAL model, evaluating performance through F1 score, accuracy (AC), and overall performance score.The overall performance score is computed as an average. The baseline SEAL model, integrating all components, serves as a reference point with robust performance. The study assesses the impact of removing individual components (core extraction, entity candidate generation, calibration, and local memory), where each removed module is replaced by a default method to ensure fair comparison. The omission of entity candidate generation affects precision, possibly due to compensatory effects in candidate selection. Calibration removal reduces output reliability, highlighting its optimization role, while the absence of local memory slightly diminishes performance stability, though less than calibration.The reliability is measured through three independent runs, referred to as the reliability experiment, to assess consistency across trials. Comparative analysis indicates a hierarchical dependency, with core extraction and entity candidate generation exerting the greatest influence due to their core functions, while calibration and local memory enhance robustness through synergy. The results affirm the interdependent nature of these components, as their individual removal consistently degrades performance, validating their collective importance to the SEAL framework.

\subsection{Low Resource Senario Study}

\begin{table}[]
\centering
\caption{Performance comparison between SEAL variants and baselines under few-shot and zero-shot settings}
\begin{tabular}{lccc}
\toprule[1pt]
                                                     & F1    & AC    & Overall \\ \hline
KB-Binder (few-shot)                                 & 32.61 & 36.82 & 34.02   \\
SEAL$_{base}$ (few-shot)                             & 41.34 & 35.39 & 39.36   \\
KB-Binder (zero-shot)                                & 21.67 & 15.96 & 19.76   \\
SEAL$_{self\_evolving}$ (zero-shot)               & 34.42 & 34.10 & 34.32   \\
\bottomrule[1pt]
\end{tabular}
\label{table:5}
\end{table}

Table~\ref{table:5} presents the results of an ablation study for zero-shot and few-shot settings, with evaluation metrics F1 score, accuracy (AC), and overall performance score. KB-Binder, in the few-shot setting, shows an F1 score of 32.61, an AC of 36.82, and an overall score of 34.02, serving as the baseline reference. SEAL-base, also in the few-shot setting, shows an F1 of 41.34, an AC of 35.39, and an overall score of 39.36, indicating superior performance compared to few-shot KB-Binder. KB-Binder lacks data in the zero-shot setting, while SEAL-self-evolving, under zero-shot, shows an F1 of 34.42, an AC of 34.10, and an overall score of 34.32, demonstrating  performance improvement through the self-improvement mechanism. Comparative analysis reveals that SEAL-base excels in the few-shot task, whereas SEAL-self-improvement shows potential enhancement in the zero-shot task, validating the efficacy of the self-improvement mechanism in data-scarce scenarios.

\subsection{Case Study}
\label{sec:casestudy}
To validate the effectiveness of the SEAL framework in complex dialogue scenarios, we conducted an analysis of a typical multi turn question answering case. This case involves identifying implicit semantic constraints, resolving cross turn coreference, and integrating user feedback. 
Successfully handling such scenarios requires dynamic dialogue state tracking and iterative refinement of the reasoning path.
Traditional methods based on static templates, such as KB Binder, often fail in such situations due to their lack of contextual awareness. 
In contrast, SEAL achieves more robust reasoning through its self-evolving mechanism, which integrates global memory, local memory, and semantic calibration.

\begin{figure}[H]
    \centering
    \includegraphics[width=1\linewidth]{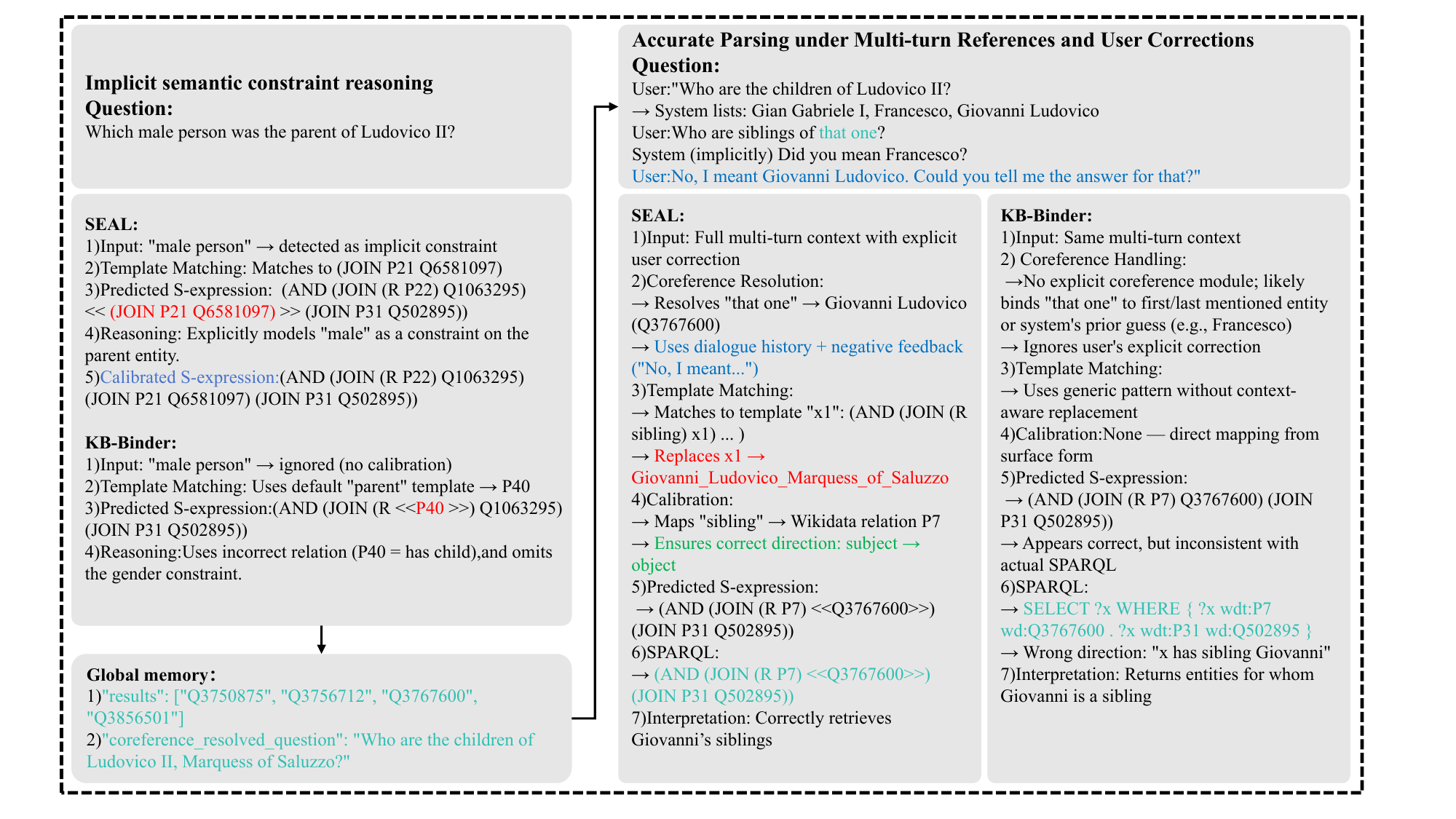}
    \caption{Comparison of SEAL and KB-Binder on a multi-turn QA example from the SPICE dataset.}
    \label{case study}
\end{figure}

As illustrated in Figure~\ref{case study}, the dialogue begins with a seemingly simple query: ``Who are the children of Ludovico II, Marquess of Saluzzo?'' However, the user's actual intent is to identify his male offspring, as all subsequent references and reasoning revolve around male heirs. This query involves two key semantic components: retrieving the ``child'' relationship and imposing an implicit constraint that the result must be a male person. KB-Binder ignores this gender constraint during initial parsing, defaulting to a generic ``parent'' template (P40) without modeling the gender restriction, which may return non-target entities such as female relatives or individuals of unknown sex, thereby contaminating the candidate set for later coreference resolution. In contrast, SEAL detects ``male person'' as an implicit constraint through calibration, explicitly models it as a type restriction (e.g., P31: Q502895), and incorporates it into the S-expression, ensuring that only the four male children, namely Gian Gabriele I, Francesco, Giovanni Ludovico, and Michele Antonio, are returned. This calibrated result is then stored in global memory, providing a clean and semantically consistent foundation for subsequent multi-turn reasoning.

In the following turn, the user asks, ``Who are siblings of that one?" The system initially infers the referent as Francesco, but the user corrects it explicitly: ``No, I meant Giovanni Ludovico." In particular, ``Giovanni Ludovico" does not appear in the user's input but was listed in the previous answer. KB-Binder lacks cross-turn memory and relies solely on surface-form entity linking, making it prone to incorrectly bind ``that one" to the first-mentioned entity or the system’s prior guess, such as Francesco, and does not respond to explicit user correction. In contrast, SEAL resolves the coreference by leveraging global memory to access the previously retrieved list of children and their corresponding QIDs, including Q3767600 for Giovanni Ludovico, thereby constraining the search space to a reliable candidate set. Local memory records the negative feedback (``No, I meant..."), and reflection enables dynamic elimination of incorrect options, confirming Giovanni Ludovico as the intended subject. Based on this accurate entity identification, SEAL generates a correctly oriented SPARQL query: {SELECT ?x WHERE \{ wd:Q3767600 wdt:P7 ?x . ?x wdt:P31 wd:Q502895 \}}, retrieving all siblings including Michele Antonio, who was never mentioned by the user. Meanwhile, KB-Binder either fails to resolve the correct entity or produces a query with reversed relation direction ({?x wdt:P7 wd:Q3767600}), resulting in incomplete or logically inconsistent results. This case demonstrates how SEAL achieves robust multi-turn understanding through a synergistic pipeline of implicit constraint calibration, memory-augmented coreference resolution, and feedback-driven refinement.

This case study demonstrates how SEAL evolves its reasoning continuously across multiple turns of dialogue. It starts with single turn semantic calibration, then coordinates global memory, local memory, and a reflection mechanism. This multilayered architecture for memory and calibration allows the system to genuinely comprehend the user's implicit intentions within the flow of conversation. 
This capability enables SEAL to outperform baselines like KB-Binder, which lack mechanisms for cross-turn context integration and feedback-driven refinement.
\section{Conclusion}
\label{sec:conclu}
By decomposing complex logical form generation into core extraction and template composition, and leveraging agent-based calibration to address syntactic and linking limitations of LLMs, SEAL's two-stage framework based on S-expression core extraction markedly enhances the accuracy of complex logical form generation. SEAL improves structure overlap and parsing success rates by 22.6\% and 22.1\%, respectively compared to KB-BINDER.
%\noindent\textbf{Future Work}

The S-expression core extraction method for semantic parsing shows robust experimental performance, yet several avenues warrant further exploration. 
Our evaluation is constrained by the fragmented nature of current KBQA benchmarks, which prevents direct comparison across diverse task formulations. Future work should establish more unified evaluation frameworks that enable fair comparison between single-turn and multi-turn KBQA systems across different knowledge bases.
To enhance scalability, the current S-expression to SPARQL transformation, constrained by specific syntactic patterns and datasets, could be generalized through versatile transformation functions validated across diverse benchmarks. Additionally, the calibration strategy, which targets non-empty query outputs, may inadvertently revise valid empty queries, reducing precision; refining the agent’s semantic understanding to accurately distinguish such cases could improve robustness. Furthermore, the LLM’s reliance on prompts for S-expression comprehension could be strengthened through targeted training on S-expression generation tasks to deepen syntactic and semantic proficiency. Finally, the computational overhead from multiple LLM invocations during core extraction could be mitigated by employing smaller models for subtasks like coreference resolution or question type classification, thereby improving efficiency.

\section*{Acknowledgments}

This work was supported in part by the Guangdong S\&T Programme (No. 2025B0101120006), the Fundamental Research Funds for the Central Universities (the Start-up Fund from Beijing Normal University, No. 310425209503), and the funding from China Telecom Research Institute (No. 26HQBYYF5024-001).



\bibliographystyle{unsrt}
\bibliography{intro+rw}
\appendix
\section{}
\label{sec:appendixA}
%%%%多轮对话案例
SPICE is a multi-turn knowledge graph question-answering dataset featuring coreference and ellipsis in dialogues. To enhance LLM’s understanding of question intent, coreference and ellipsis are resolved by providing the LLM with historical user-system interactions, focusing on the user’s final question, and reformulating them into a complete, standalone query. This resolved question replaces the original incomplete one for subsequent processing.

\framebox[\textwidth]{%
\begin{minipage}{\dimexpr\textwidth-2\fboxsep-2\fboxrule}

Input:
\begin{itemize}
    \item Which people emerged victorious in La Madrid Challenge by La\\
          Vuelta 2016 and La Madrid Challenge by La Vuelta 2015?
    \item Giorgia Bronzini, Shelley Olds, Kirsten Wild
    \item Which television programs are that person a screenwriter of?
    \item Did you mean Giorgia Bronzini?
    \item No, I meant Shelley Olds. Could you tell me the answer for that?
\end{itemize}

\% [ENTITIES]\\
\% La Madrid Challenge by La Vuelta 2016:\\
\% cycling\_race\_class\_defined\_by\_the\_International\_Cycling\_Union;\\
\% La Madrid Challenge by La Vuelta 2015:\\
\% cycling\_race\_class\_defined\_by\_the\_International\_Cycling\_Union;\\
\% Giorgia Bronzini: common\_name;\\
\% Shelley Olds: common\_name;\\
\% Kirsten Wild: common\_name

Output:
\begin{itemize}
    \item Which television programs is Shelley Olds a screenwriter of?
\end{itemize}

\end{minipage}}
\section{}

\label{sec:appendixb}
%%s表达核心示例
This article presents the `S-expression core' a concise subset of S-expressions limited to JOIN, R, AND, VALUES, and IS\_TRUE functions. In natural language queries, S-expression cores denote objects of inquiry or comparison. In SPARQL queries, they map to graph patterns in the WHERE\{\ldots\} clause. Training data S-expression cores are derived from these patterns. Figure 5 shows these mappings with examples of questions and S-expression cores.

\framebox[\textwidth]{%
\begin{minipage}{\dimexpr\textwidth-2\fboxsep-2\fboxrule}

Question: Which male person was the parent of Ludovico II, Marquess of Saluzzo?

S-expression: (AND (JOIN (R father) Ludovico\_II,\_Marquess\_of\_Saluzzo) (JOIN instance\_of common\_name))

S-expression Core: (AND (JOIN (R father) Ludovico\_II,\_Marquess\_of\_Saluzzo) (JOIN instance\_of common\_name))

Question: How many people have Gian Gabriele I of Saluzzo as their sibling?

S-expression: (COUNT (AND (JOIN (R brother) Gian\_Gabriele\_I\_of\_Saluzzo) (JOIN instance\_of common\_name)))

S-expression Core: (AND (JOIN (R brother) Gian\_Gabriele\_I\_of\_Saluzzo) (JOIN instance\_of common\_name))

\end{minipage}
}

%%%S-表达式核心与问题和 SPARQL 的对应关系
% \begin{figure}[ht]
%     \centering
%     \includegraphics[width=0.90\linewidth]{img/image.svg}
%     \caption{Sexpression}
%     \label{figure:S_expression}
% \end{figure}%表达式心模式

\begin{figure}[ht]
    \centering
    \includegraphics[width=1\linewidth]{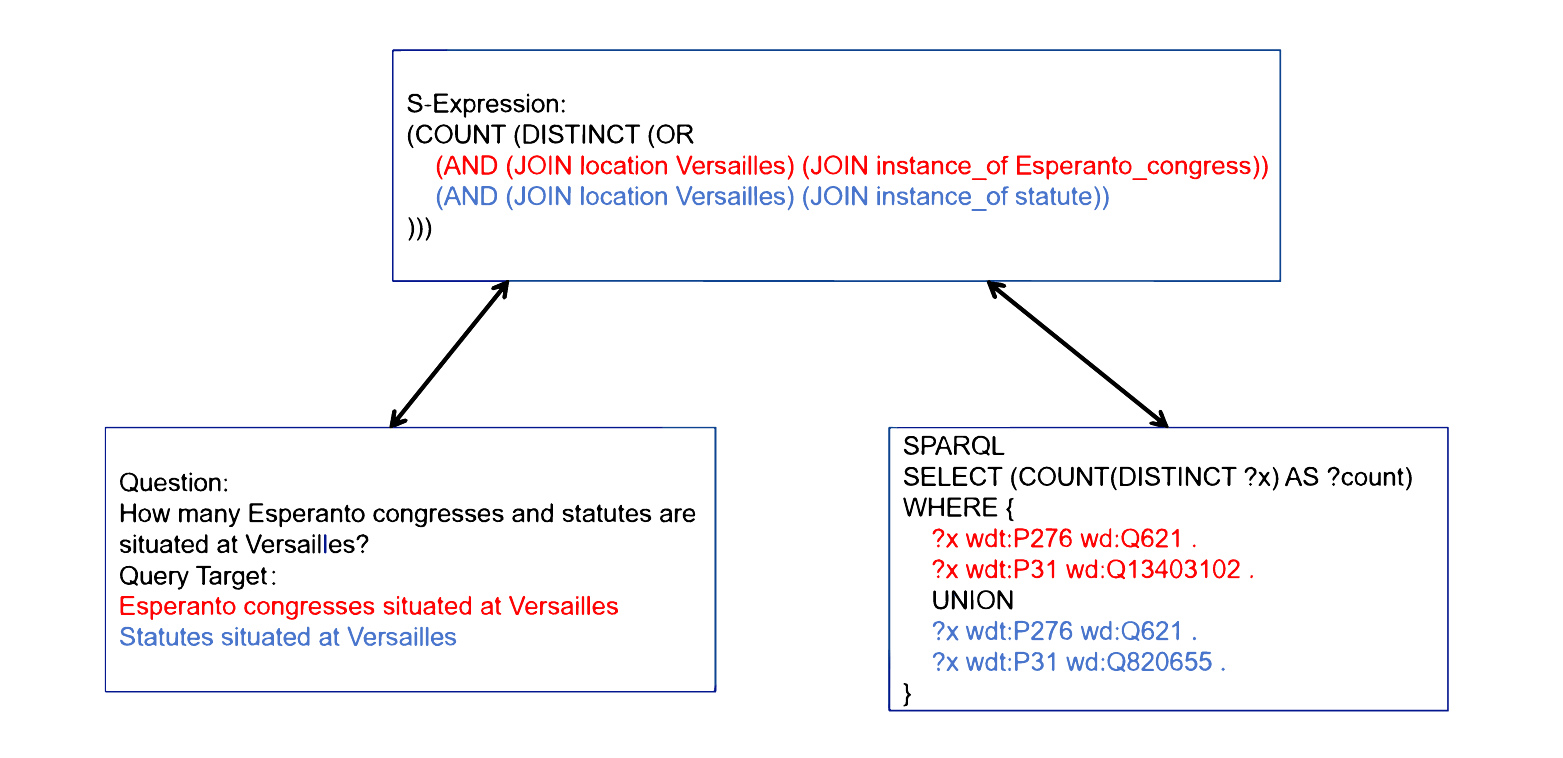}
    \caption{S-expression}
    \label{fig:Sexpression}
\end{figure}
\section{}
%%%S-表达式核心模式

\label{sec:appendixc}
After the LLM generates the S-expression core, an Agent is required to calibrate the generated core, so we summarize the common patterns of S-expression cores as follows.

\begin{center}
\begin{tabular}{p{10cm}}
\hline
\textbf{S-Core pattern of expression} \\
\hline
1. (IS\_TRUE x x x) \\
2. (JOIN x x) \\
3. (JOIN (R x) x) \\
4. (AND (JOIN x x) (JOIN x x)) \\
5. (AND (JOIN x (VALUES x ...)) (JOIN x x)) \\
6. (AND (JOIN (R x) x) (JOIN x x)) \\
7. (AND (JOIN (R x) (VALUES x ...)) (JOIN x x)) \\
8. (AND (JOIN x (JOIN x x)) (JOIN x x)) \\
9. (AND (JOIN (R x) (JOIN x x)) (JOIN x x)) \\
10. (AND (JOIN x x) (JOIN x x) (JOIN x x)) \\
11. (AND (JOIN (R x) x) (JOIN (R x) x) (JOIN x x)) \\
12. (AND (JOIN (R x) x) (JOIN x x) (JOIN (R x) x)) \\
\hline
\end{tabular}
\end{center}

%%%S-表达式模板与对应的问题类型
% \begin{figure}[ht]
%     \centering
%     \includegraphics[width=1\linewidth]{img/0e18acaaf7cd79ace5aebf4a77dce1a.pdf}
%     \caption{2}
%     \label{figure:2}
% \end{figure}
\section{}

\label{sec:appendixd}
During the organizing of S expression templates, we identified an error in the SPICE dataset. For questions involving equality relations, the corresponding SPARQL queries often lack the necessary FILTER component to screen for equality, resulting in incomplete S-expressions. For example, the correct template should have the compare function set to EQ, but the actual template is (GROUP \_COUNT x1).  

The following describes the two primary steps for generating complete S-expressions from templates.

\begin{tabular}{|l|l|}
\hline
\textbf{Operation} & \textbf{S-based Query Syntax} \\
\hline
single & x1 \\
& (OR x1 x2) \\
& (DISTINCT x1) \\
& (DIFF x1 x2) \\
& *(GROUP\_COUNT x1) \\
& *(GROUP\_SUM (GROUP\_COUNT x1) (GROUP\_COUNT x2)) \\
\hline
verify & (ALL x1) \\
& (ALL x1 x2) \\
& (ALL x1 x2 x3) \\
& ... \\
\hline
count & (COUNT x1) \\
& (COUNT (DISTINCT x1)) \\
& (COUNT (DISTINCT (OR x1 x2))) \\
& *(COUNT (GROUP\_COUNT x1)) \\
& *(COUNT (GROUP\_SUM (GROUP\_COUNT x1))) \\
& (GROUP\_COUNT x2) \\
\hline
compare & (compare (GROUP\_COUNT x1) number) \\
& (compare (GROUP\_COUNT x1 x2)) \\
& (compare (GROUP\_SUM (GROUP\_COUNT x1))) \\
& (GROUP\_COUNT x2) number) \\
& (compare (GROUP\_SUM (GROUP\_COUNT x1))) \\
& (GROUP\_COUNT x2) (OR x3 x4)) \\
\hline
compare\_and\_count & (COUNT (compare (GROUP\_COUNT x1) number)) \\
& (COUNT (compare (GROUP\_COUNT x1) x2)) \\
& (COUNT (compare (GROUP\_SUM))) \\
& (GROUP\_COUNT x1 (GROUP\_COUNT x2) number)) \\
& (COUNT (compare (GROUP\_SUM))) \\
& (GROUP\_COUNT x1 (GROUP\_COUNT x2) (OR x3 x4))) \\
\hline
optimize & (optimize (GROUP\_COUNT x1)) \\
& (optimize (GROUP\_SUM (GROUP\_COUNT x1))) \\
& (GROUP\_COUNT x2)) \\
\hline
\end{tabular}

\section{}

\label{sec:appendixe}
%%会根据语法生成候选以外的正确模板，
Notably, while we refer to the process as template ``selection'', the model retains adaptive flexibility, often generating correct templates beyond the candidate set based on syntactic context, as illustrated in the example below.

\framebox[\textwidth]{%
\begin{minipage}{\dimexpr\textwidth-2\fboxsep-2\fboxrule}

Correct S-expression:\\
(AND (JOIN (R field\_of\_this\_occupation) (VALUES assistant\_coach\\
\hspace*{0.5cm} association\_football\_manager association\_football\_player)) (JOIN\\
\hspace*{0.5cm} instance\_of sport))

Predicted S-expression Core:\\
(AND (JOIN (R field\_of\_this\_occupation) assistant\_coach) (JOIN\\
\hspace*{0.5cm} instance\_of sport)),\\
(AND (JOIN (R field\_of\_this\_occupation) association\_football\_manager) (\\
\hspace*{0.5cm} JOIN instance\_of sport)),\\
(AND (JOIN (R field\_of\_this\_occupation) association\_football\_player) (\\
\hspace*{0.5cm} JOIN instance\_of sport))

Generated Template:\\
(OR (OR x1 x2) x3) --- Not in the template library

\end{minipage}
}

\section{}

\label{sec:appendixf}
%%%%模板选择与替换方案生成
We provide the LLM with the question, candidate templates (determined by the predicted question type), and calibrated S-expression core variants, enabling it to select an appropriate template and generate a substitution plan. The final S-expression is obtained by replacing placeholders in the selected template with the generated plan. An example of input and output is provided below.

\framebox[\textwidth]{%
\begin{minipage}{\dimexpr\textwidth-2\fboxsep-2\fboxrule}
\small % Reduce font size
Input

Question:
How many administrative territories are the narrative locations of at least 840 applications or works of art?

Candidate Templates (Predicted question type: compare\_and\_count):
\begin{itemize}
    \item (COUNT (compare (GROUP\_COUNT x1) number)),
    \item (COUNT (compare (GROUP\_COUNT x1) x2)),
    \item (COUNT (compare (GROUP\_SUM (GROUP\_COUNT x1) (GROUP\_COUNT x2)) number)),
    \item (COUNT (compare (GROUP\_SUM (GROUP\_COUNT x1) (GROUP\_COUNT x2)) (OR x3 x4)))
\end{itemize}

Candidate S-expression Core:
\begin{itemize}
    \item (AND (JOIN (R narrative\_location) (JOIN instance\_of application)) (JOIN instance\_of administrative\_territorial\_entity)),
    \item (AND (JOIN (R narrative\_location) (JOIN instance\_of work\_of\_art)) (JOIN instance\_of administrative\_territorial\_entity))
\end{itemize}

Output

Template:
(COUNT (compare (GROUP\_SUM (GROUP\_COUNT x1) (GROUP\_COUNT x2)) number))

Replacement Plan:
\{\\
"variables": \{\\
\hspace*{1cm}"x2": "(AND (JOIN (R narrative\_location) (JOIN instance\_of work\_of\_art)) (JOIN instance\_of administrative\_territorial\_entity))",\\
\hspace*{1cm}"x1": "(AND (JOIN (R narrative\_location) (JOIN instance\_of application)) (JOIN instance\_of administrative\_territorial\_entity))"\\
\},\\
"constants": \{\\
\hspace*{1cm}"number": 840\\
\},\\
"functions": \{\\
\hspace*{1cm}"compare": "GE"\\
\}\\
\}

\end{minipage}
}

% \appendix

\section{Implementation Details of S-expression Core Calibration}
\label{app:calibration_details}

To ensure robustness against syntactic errors and semantic hallucinations in LLM-generated S-expressions, we implemented a rigorous calibration pipeline. This pipeline consists of five key operations: (1) Parenthesis Matching, (2) Entity Identification \& Replacement, (3) Relation Identification \& Replacement, (4) Type Constraint Enforcement, and (5) Expression Variant Generation.

\subsection{Recursive Syntactic Reconstruction Rules}
LLMs often generate S-expressions with unbalanced parentheses or incorrect nesting structures. To address this, we employ a \textbf{Recursive Reconstruction} strategy that rebuilds the syntax tree from flat tokens. The procedure is as follows:

\begin{enumerate}
    \item \textbf{Tokenization \& Cleaning:} The input S-expression is first stripped of all existing parentheses to eliminate structural noise and split into a sequence of atomic tokens.
    \item \textbf{Arity-based Parsing:} A recursive parser processes the token stream. For each function token encountered, the parser recursively consumes a specific number of subsequent arguments based on the function's definition:
    \begin{itemize}
        \item \textbf{IS\_TRUE:} Consumes 3 arguments (Subject, Predicate, Object).
        \item \textbf{JOIN:} Consumes 2 arguments (Relation/Entity, Entity).
        \item \textbf{R:} Consumes 1 argument (Relation).
        \item \textbf{AND:} Consumes a variable number of arguments until the end of the scope.
    \end{itemize}
    \item \textbf{Reconstruction:} The parser wraps the consumed arguments in correctly balanced parentheses (e.g., \texttt{(Function Arg1 Arg2)}) before returning the substructure. This method guarantees that the output S-expression is syntactically valid regardless of the original depth of nesting or missing brackets.
\end{enumerate}

\subsection{Execution-guided Semantic Alignment Algorithm}
After syntactic correction, the expression contains surface names (e.g., ``sorbitol\_dehydrogenase'') that need to be grounded to Knowledge Graph (KG) identifiers. We employ an execution-guided search algorithm:

\begin{enumerate}
    \item \textbf{Candidate Retrieval:} For each surface form token (entity or relation), we retrieve the top-$k$ candidates using a dense retriever based on semantic similarity.
    \item \textbf{Constraint Injection:} We detect implicit constraints such as \texttt{instance\_of} and strictly map them to the Wikidata property \texttt{P31} (``instance of'') to filter entity candidates by type.
    \item \textbf{Cartesian Product Generation:} We generate a set of candidate S-expressions by computing the Cartesian product of all possible entity and relation candidates.
    \item \textbf{Structural Variation:} To handle potential directionality errors in relations, we automatically generate variants with reversed relations (e.g., swapping Subject and Object).
    \item \textbf{Execution Verification:} Each candidate variant is translated into SPARQL and executed against the KG. The search terminates at the first variant that returns a \textbf{non-empty result}, which is then selected as the final calibrated core.
\end{enumerate}

\subsection{Qualitative Case Study}
Table \ref{tab:calibration_case} demonstrates the robustness of the calibration module. The LLM initially predicted an expression with incorrect relation surface forms (e.g., \texttt{involved\_in}) and unlinked entities. The calibration module successfully corrected the syntax, mapped ambiguous relations to specific properties (e.g., mapping \texttt{involved\_in} to \texttt{P703} based on the context of the entity), and enforced type constraints.

\begin{table}[h]
    \centering
    \caption{Case Study of Calibration Process}
    \label{tab:calibration_case}
    \renewcommand{\arraystretch}{1.5}
    \begin{tabular}{lp{10cm}}
        \toprule
        \textbf{Stage} & \textbf{S-expression Content / Note} \\
        \midrule
        \textbf{Prediction} & \texttt{(AND (JOIN (R involved\_in) sorbitol\_dehydrogenase\_Lmo2664) (JOIN instance\_of metabolic\_process))} \\
        \textit{Issue} & Contains surface forms; \texttt{involved\_in} is ambiguous; potential parenthesis errors. \\
        \midrule
        \textbf{Calibrated (IDs)} & \texttt{(AND (JOIN (R P703) Q24251808) (JOIN P31 Q631147))} \\
        \textit{Result} & \textbf{Executable.} Entities and relations are grounded to Wikidata IDs. \\
        \midrule
        \textbf{Interpretation} & \texttt{(AND (JOIN (R found\_in\_taxon) sorbitol\_dehydrogenase...lmo2664) (JOIN instance\_of infraspecific\_name))} \\
        \textit{Logic} & The logic is refined: ``Find entities found in taxon lmo2664 that are infraspecific names''. \\
        \bottomrule
    \end{tabular}
\end{table}

\section{Robustness Analysis on Out-of-Distribution Data}
\label{app:ood_analysis}

To evaluate the 'retraining-free adaptation' capability of the self-evolving mechanism, we conducted additional ablation studies on a constructed Out-of-Distribution (OOD) test set. The construction process was designed to simulate real-world scenarios where user queries often deviate from training patterns. Specifically, we utilized the 3,971 samples from the SPICE dataset (used in the main experiment) as the original knowledge source. While strictly maintaining the verified entity and relation combinations from the database to ensure executability, we introduced significant perturbations to the query surface forms. We focused on maximizing the distributional distance in critical features, such as \textit{lexical distribution} (substituting common terms with rare synonyms or paraphrases) and \textit{question structure} (transforming simple canonical questions into complex syntactic structures with nested clauses). Based on these principles, we generated a challenging subset of 200 data samples that are semantically equivalent to the original knowledge but syntactically and lexically distinct.
We evaluated our method on this newly generated OOD dataset to verify its generalization performance. The experimental results are summarized in Table \ref{tab:ood_results}. As anticipated, the introduction of substantial distributional shifts and increased query complexity led to a performance decrement compared to the original In-Distribution (ID) test set. Specifically, unseen linguistic patterns pose a greater challenge for the initial parsing stage. However, the performance decline remains within a reasonable and acceptable range. This resilience indicates that despite the lack of direct supervision on these new patterns, the self-evolving module effectively mitigates the gap by dynamically calibrating the S-expressions through interaction with the Knowledge Graph. These results confirm that SEAL has strong robustness and can adapt to unseen domains without the need for parameter retraining.

\begin{table}[htbp]
\centering
\caption{SEAL performance over newly constructed dataset}
\label{tab:ood_results}
\begin{tabular}{lccc}
\toprule
\textbf{Question Type} & \textbf{Instances} & \textbf{Macro-F1} & \textbf{Accuracy} \\
\midrule
Logical Reasoning (All) & 24 & 0.5863 & --- \\
Quantitative Reasoning (All) & 26 & 0.4153 & --- \\
Comparative Reasoning (All) & 17 & 0.3529 & --- \\
Simple Question (Coreferenced) & 21 & 0.5429 & --- \\
Simple Question (Direct) & 28 & 0.6429 & --- \\
Simple Question (Ellipsis) & 18 & 0.5378 & --- \\
Verification (Boolean) (All) & 25 & --- & 0.8800 \\
Quantitative Reasoning (Count) (All) & 25 & --- & 0.8000 \\
Comparative Reasoning (Count) (All) & 16 & --- & 0.1875 \\
\midrule
Overall & 200 & \multicolumn{2}{c}{0.5748} \\
\bottomrule
\end{tabular}
\end{table}

\section{Generalization Across Different Foundation Models}
\label{app:llm_robustness}

To further assess the robustness and model-agnostic capability of the SEAL framework, we conducted additional ablation studies by replacing the backbone Large Language Model. Specifically, we employed the \textbf{Qwen3-Instruct} model (checkpoint: \textit{Qwen3-235B-A22B-Instruct-2507}) to replace the original backbone used in the main experiments. This experiment aims to verify whether SEAL can maintain high performance and effective self-evolution when the underlying generator's scale and pre-training distribution vary.

\subsection{Experimental Results and Analysis}
We evaluated the model on the full test set using the same metrics as the main experiment. The detailed breakdown of performance by question type is presented in Table \ref{tab:qwen_results}.

\begin{table}[htbp]
\centering
\caption{Performance of SEAL using \textit{Qwen3-Instruct} as the backbone LLM across different question types.}
\label{tab:qwen_results}
\footnotesize % 减少字体大小
\begin{tabular}{lccccc}
\toprule
\textbf{Question Type} & \textbf{Instances} & \textbf{Precision} & \textbf{Recall} & \textbf{F1} & \textbf{Accuracy} \\
\midrule
Logical Reasoning & 321 & 0.9882 & 0.6494 & 0.7838 & --- \\
Quantitative Reasoning & 213 & 0.7307 & 0.3634 & 0.4854 & --- \\
Comparative Reasoning & 286 & 0.8383 & 0.1957 & 0.3173 & --- \\
Simple Question (Coref) & 606 & 0.4373 & 0.2395 & 0.3095 & --- \\
Simple Question (Direct) & 640 & 0.9696 & 0.4544 & 0.6188 & --- \\
Simple Question (Ellipsis) & 161 & 0.9789 & 0.5054 & 0.6667 & --- \\
Verification (Boolean) & 355 & --- & --- & --- & 0.4958 \\
Quantitative Reasoning (Count) & 481 & --- & --- & --- & 0.4304 \\
Comparative Reasoning (Count) & 297 & --- & --- & --- & 0.1380 \\
\bottomrule
\end{tabular}
\end{table}

As indicated in Table \ref{tab:qwen_results}, the SEAL framework achieves consistent and competitive results with the Qwen3 backbone. While the overall Accuracy  shows a variation compared to the main experimental results, this is primarily attributed to the intrinsic differences in code generation preferences between varying foundation models. Some LLMs may exhibit weaker initial alignment with the specific S-expression syntax required for KBQA tasks. However, it is crucial to note that the \textit{self-evolving mechanism} within SEAL successfully mitigated these gaps by calibrating the initial predictions through interaction with the knowledge graph. The results confirm that SEAL is robust across different model architectures and scales, effectively enhancing the capabilities of base models regardless of their initial proficiency in semantic parsing.

\section{API resource consumption}
\label{app:api_cons}
We tracked the average number of LLM API calls. As shown in Table~\ref{tab:api_consumption}, our system made a total of 14,303 API requests and consumed 23,537,434 tokens on 3360 queries. This corresponds to an average of approximately 3.56 API calls per question, with the S-expression prediction module accounting for the majority of token consumption.

\begin{table}[ht]
\centering
\caption{API resource consumption across different modules of SEAL during evaluation.}
\label{tab:api_consumption}
\begin{tabular}{lrr}
\toprule
\textbf{Module} & \textbf{Requests} & \textbf{Tokens} \\
\midrule
Coreference Resolution      & 7,573   & 3,054,479 \\
Extract S-expression Cores  & 155     & 1,933,432 \\
Predict Question Type       & 3,360   & 3,100,668 \\
Predict S-expression        & 3,215   & 15,448,855 \\
\midrule
\textbf{Overall}            & \textbf{14,303} & \textbf{23,537,434} \\
\bottomrule
\end{tabular}
\end{table}

\section{Quantitative Analysis of out of Template Scenario}
\label{app:template_analysis}

To quantify the coverage of our template mechanism and verify the generalization capability of the LLM when templates are unavailable or mismatched, we conducted a statistical analysis of the \textit{out-of-template generation accuracy}. We conducted an experiment on unmatched template condition. The statistical results and corresponding performance metrics of unmatched templates scenario are presented in Table \ref{tab:template_stats}.

\begin{table}[htbp]
\centering
\caption{Performance of SEAL in unmatch templates scenario}
\label{tab:template_stats}
\footnotesize % 缩小字体（比 \footnotesize 稍大，可读性更好）
\setlength{\tabcolsep}{4pt} % 默认是 6pt，减小列间距
\begin{tabular}{@{}lcccccc@{}}
\toprule
\textbf{Question Type} & \textbf{Instances} & \textbf{Precision} & \textbf{Recall} & \textbf{F1-Score} & \textbf{Macro-F1} & \textbf{Accuracy} \\
\midrule
Simple Question (Direct) & 108 & 0.7332 & 0.2698 & 0.3944 & 0.3288 & --- \\
Simple Question (Coreferenced) & 199 & 0.6290 & 0.5120 & 0.5645 & 0.3542 & --- \\
Comparative Reasoning (Count) (All) & 167 & --- & --- & --- & --- & 0.0240 \\
Quantitative Reasoning (Count) (All) & 407 & --- & --- & --- & --- & 0.5700 \\
Logical Reasoning (All) & 177 & 0.9670 & 0.6805 & 0.7989 & 0.5024 & --- \\
Verification (Boolean) (All) & 4 & --- & --- & --- & --- & 0.0000 \\
Simple Question (Ellipsis) & 26 & 0.0651 & 0.3590 & 0.1102 & 0.2063 & --- \\
Quantitative Reasoning (All) & 95 & 0.1288 & 0.0340 & 0.0538 & 0.2056 & --- \\
\bottomrule
\end{tabular}
\end{table}
The template hit rate is 45.5\% (1724/3791), and as shown in Table \ref{tab:template_stats}, SEAL method serving as effective scaffolding for complex semantic parsing.  Crucially, the SEAL framework maintains a good level of accuracy. This indicates that our method does not rigidly depend on templates. Instead, when templates are absent or imperfect, the \textit{self-evolving module} effectively compensates by dynamically calibrating the S-expressions. This validates that SEAL possesses strong generalization capabilities, using templates to accelerate reasoning when possible, while retaining the robustness to handle novel query structures independently.

\end{document}